\documentclass{article} %
\usepackage{iclr2025_conference,times}

\usepackage{amsmath,amsfonts,bm}

\def\eqref#1{equation~\ref{#1}}

\def\1{\bm{1}}

\DeclareMathAlphabet{\mathsfit}{\encodingdefault}{\sfdefault}{m}{sl}
\SetMathAlphabet{\mathsfit}{bold}{\encodingdefault}{\sfdefault}{bx}{n}

\newcommand{\E}{\mathbb{E}}

\usepackage{hyperref}
\usepackage{url}
\usepackage{soul}
\usepackage{natbib}

\usepackage[utf8]{inputenc} %
\usepackage[T1]{fontenc}    %
\usepackage{booktabs}       %
\usepackage{amsfonts}       %
\usepackage{nicefrac}       %
\usepackage{microtype}      %
\usepackage{xcolor}         %

\usepackage{graphicx}
\usepackage{wrapfig}
\usepackage{caption}

\usepackage{algorithm2e}
\RestyleAlgo{ruled}
\makeatletter
\renewcommand{\@algocf@capt@plain}{above}%
\makeatother

\usepackage{amsmath, amssymb}
\usepackage{cleveref}

\usepackage{amsmath,amsfonts,amsthm,bm, bbm}

\newtheorem{lemma}{Lemma}

\title{Deconstructing What Makes a Good Optimizer for Autoregressive Language Models}

\author{
Rosie Zhao\thanks{Equal contribution, randomized author ordering. Correspondence to \texttt{rosiezhao@g.harvard.edu}.}\\
Harvard University
\And
Depen Morwani$^*$\\
Harvard University
\And
David Brandfonbrener$^*$\\
Kempner Institute at Harvard University
\And
Nikhil Vyas$^*$\\
Harvard University
\And 
Sham Kakade\\
Kempner Institute at Harvard University
}

\iclrfinalcopy %
\begin{document}

\maketitle

\begin{abstract}
Training language models becomes increasingly expensive with scale, prompting numerous attempts to improve optimization efficiency. Despite these efforts, the Adam optimizer remains the most widely used, due to a prevailing view that it is the most effective approach. We aim to compare several optimization algorithms, including SGD, Adafactor, Adam, Lion, and Sophia in the context of autoregressive language modeling across a range of model sizes, hyperparameters, and architecture variants. Our findings indicate that, except for SGD, these algorithms all perform comparably both in their optimal performance and also in terms of how they fare across a wide range of hyperparameter choices. Our results suggest to practitioners that the choice of optimizer can be guided by practical considerations like memory constraints and ease of implementation, as no single algorithm emerged as a clear winner in terms of performance or stability to hyperparameter misspecification. Given our findings, we further dissect these approaches, examining two simplified versions of Adam: a) signed momentum (Signum)  which we see recovers both the performance and hyperparameter stability of Adam and b) Adalayer, a layerwise variant of Adam which we introduce to study the impact on Adam's preconditioning for different layers of the network. Examining Adalayer leads us to the conclusion that, perhaps surprisingly, adaptivity on \textit{both} the last layer and LayerNorm parameters in particular are necessary for retaining performance and stability to learning rate. %

\end{abstract}

\section{Introduction}

As language model architectures increase in scale, pretraining becomes more expensive. In response, numerous efforts have been made to design efficient optimizers to mitigate these costs, and yet Adam \citep{kingma15} remains the primary optimizer used for training language models. This persistent preference for Adam is rooted in an underlying belief that Adam generally outperforms alternative optimization algorithms.
Although newly proposed optimizers run ablations to demonstrate superior performance to Adam for select architectures and tasks \citep{liu2024sophia, chen2023symbolic}, there is no consensus among the literature about the relative performance of these optimizers. In fact, to the best of our knowledge, \citet{kaddour2024no} is the only work comparing these optimizers but in the context of masked language modeling and at a single model scale.

\begin{figure}[t]
    \centering
    \includegraphics[width=0.85\textwidth]{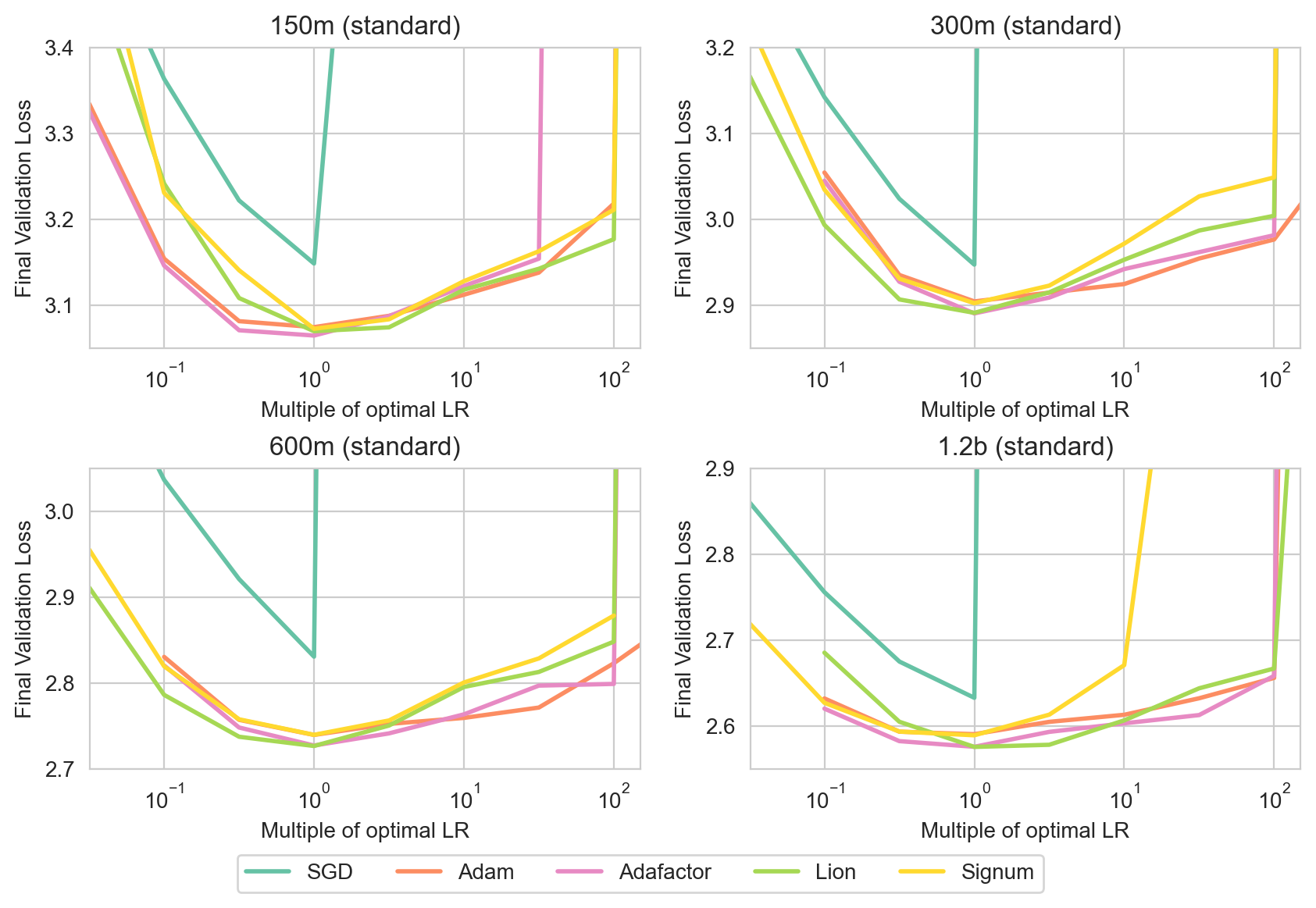}
    \caption[a]{Final validation loss when training language models with 150m, 300m, 600m, and 1.2b parameters, sweeping across learning rates for five standard optimizers (SGD, Adam, Adafactor\footnotemark{}, Lion, and Signum). Plots have been shifted to align the optimal learning rates for each optimizer. Except for SGD, other optimizers seem comparable in their optimal performance and stability with respect to learning rate tuning.}
    \label{fig:main}
\end{figure}

In this work, we aim to rigorously evaluate whether the widespread reliance on Adam is justified and to further explore the characteristics that make adaptive optimizers particularly effective for training autoregressive language models. As such, we perform a comprehensive sweep across different optimizers, hyperparameters, architectures, and scale. Along with looking at optimal performance, we argue that due to the difficulty of hyperparameter tuning with increasing scale \citep{Yang21}, the \emph{stability} of performance with respect to hyperparameter choices is equally important. Prior work has explored the learning rate stability of Adam \citep{wortsman2024smallscale}. We extend this investigation to include the stability of multiple optimizers with respect to various hyperparameter choices. Surprisingly, we find that multiple optimizers introduced in the literature after Adam---such as Lion~\citep{chen2023symbolic} and Adafactor (with momentum)~\citep{ShazeerS18,Zhai0HB22}---demonstrate robustness comparable to Adam and significantly superior to SGD. Figure \ref{fig:main} illustrates the remarkable similarity in performance and robustness of these optimizers across different learning rates and across multiple model scales (150m, 300m, 600m, and 1.2b parameters). This challenges the prevailing notion that Adam should be the default optimizer, where we see no single algorithm emerged as a clear winner in terms of performance or hyperparameter stability.

\footnotetext{Our implementation of Adafactor adds back momentum as described in Section~\ref{subsec:setup}.}

Following our initial ablations, we wish to identify the essential components of these optimizers that facilitate performance and stability. Thus, we conduct a series of investigations of simplified versions of these algorithms. We study signed momentum (Signum) \citep{pmlr-v80-bernstein18a, bernstein2018signsgd}, a special case of Lion. Prior works have also studied its similarities to Adam~\citep{balles2018dissecting}. %
We find that Signum also recovers the stability and performance exhibited by Adam. This finding aligns with recent work~\citep{kunstner2023noise}, suggesting that the primary distinction between SGD and Adam is driven by Adam's resemblance to signSGD.

To further understand the role of preconditioning on various network parameters, we study Adalayer, which performs preconditioning on a \emph{per-layer} basis. We empirically demonstrate that this variant nearly recovers the stability and performance of the other optimizers in previous ablations. Through empirical studies of Adalayer and its variants, we show that adapting the parameters of the last layer and LayerNorm parameters in a transformer is necessary to achieve stability and performance.

\textbf{Main contributions}
\begin{itemize}
    \item We empirically study the stability to hyperparameters of various optimization algorithms including SGD, Adam, Lion and Adafactor, showing that with the exception of SGD, these optimizers are comparable in terms of both performance and hyperparameter stability. This holds across multiple scales (150m, 300m, 600m, 1.2b) and across two transformer architecture variants (Section \ref{sec:ablations}).
    \item To dive into the reasons behind the stability and performance of these algorithms, we empirically examine signed momentum (Signum), and show that it recovers their stability and performance (Section \ref{sec:signsgd}).
    \item We study a coarser variant of Adam called Adalayer, that does per-layer  preconditioning and recovers much of the stability and performance exhibited by Adam (Section \ref{subsec:lastlayer_correction}).
    \item Through an empirical study of Adalayer and its variants, we study the importance of adaptivity with respect to different layers of the network. (Section \ref{subsec:sgd+novograd}).
\end{itemize}

\section{Comparing Optimizers Across Hyperparameters, Architectures and Scale}
\label{sec:ablations}

\subsection{Methodology}

To conduct our experiments, we start with hyperparameters recommended by previous work (e.g., $\beta_1 = 0.9$). We initially perform a learning rate sweep to identify the optimal learning rate. After determining the optimal learning rate for each algorithm, we conduct one-dimensional sweeps for each of the other hyperparameters.

A limitation of this methodology is the potential neglect of "higher-dimensional" interactions between hyperparameters. 
This is an important direction for future work, but beyond the computational budget of this project. 
For example, some parameters like batch size and learning rate indeed are likely to exhibit 2D interactions~\citep{shallue19,porian2024resolving}.
However, we argue that the 1D sweeps provide a tractable methodology that gives us useful signal about the hyperparameter stability of a variety of algorithms around the parameters that are common in practice.

\subsection{Setup}
\label{subsec:setup}

We train decoder-only language models on C4 tokenized with the T5 tokenizer \citep{raffel2020exploring} and report results in terms of validation loss. Due to the computational resources required to sweep over numerous hyperparameters, optimizers, and scales, we restrict the scope of our work to autoregressive language models, leaving other domains and language model architectures as future work. As we discussed in the introduction, we argue that it is best to evaluate algorithms both in terms of the loss achieved by the best hyperparameters (performance) as well as the robustness across values of the hyperparameters (stability).

In this section we first present the results of sweeps across the two most sensitive hyperparameters: learning rate and momentum ($\beta_1$). We sweep across five algorithms: Adam, Adafactor, Lion, Signum, and SGD. Further ablations of weight decay, warmup, $ \beta_2$, and $ \epsilon$ for the 150m standard model can be found in \Cref{app:ablations}. Before diving into the results, we provide some more details about the setup. 

\paragraph{Algorithms.} We use the standard Pytorch implementation of AdamW \citep{paszke2019pytorch}, the timm implementation of SGDW \citep{rw2019timm}, and the OLMo implementation of Lion \citep{groeneveld2024olmo}. Following~\cite{Zhai0HB22} we implement ourselves a modified version of Adafactor which maintains the factored estimates of second moments but \textbf{has momentum} i.e. it is equivalent to Adam with factored second moment estimates. Since Signum is equivalent to Lion with $\beta_1 = \beta_2$ we reuse the OLMo implementation of Lion \citep{groeneveld2024olmo} for it. We conducted experiments with the Sophia optimizer~\citep{liu2024sophia} in Appendix~\ref{app:sophia}. However, since it does not outperform Signum (which can be achieved by setting $\rho = 0$ in Sophia), we did not include it in other plots.

\paragraph{Models.} We start from the OLMo codebase \citep{groeneveld2024olmo} and train decoder-only transformer models of four sizes: 150m, 300m, 600m, and 1.2b, where the parameter count refers to non-embedding parameters. The models have widths of 1024, 1024, and 1408 and depths of 12, 24, 24. The MLP hidden dimension is 4x of the width. The activation function is GeLU \citep{hendrycks2016gaussian}. We use RoPE positional encodings \citep{su2024roformer}. Attention heads are always dimension 64. We use PyTorch default LayerNorm. Following \citet{wortsman2024smallscale} we do not learn biases for the linear layers or LayerNorms. We train in mixed precision with bfloat16.

\paragraph{Training variants.} We note that \citet{wortsman2024smallscale} observe that QK LayerNorm \citep{dehghani2023scaling} and z-loss \citep{chowdhery2023palm} can have substantial effects on the stability of model training. As such, we consider two variants in our experiments: \textbf{standard} which refers to a model with QK LayerNorms and z-loss with coefficient 1e-4, and \textbf{no QK norm or z-loss} which refers to the same model without the QK norm layers or the z-loss. 

\paragraph{Token counts.} For all models, we use a batch size of 256 and sequence length of 512 (as in \citet{wortsman2024smallscale}). We default to training models for the approximately ``chinchilla optimal'' number of tokens that is $ \approx$20 times the number of parameters. Explicitly, this means for the 150m models we train for 25k steps or $\approx$3.3b tokens. The 300m models are trained for 50k steps, the 600m models are trained for 100k steps, the 1.2b models are trained for 200k steps, and the 150m-long models are also trained for 100k steps. 

\paragraph{Other hyperparameters.} We default to using 0 weight decay. We default to using a learning rate schedule with 10\% of the training steps for warmup and then cosine decay with a minimum that is 10\% of the maximum learning rate. We default to $ \beta_2 = 0.95$ and $ \epsilon = 1\text{e-}15$ following \citet{wortsman2024smallscale}. These parameters are ablated in \Cref{app:ablations}.

\subsection{Sweeping learning rates}

\begin{figure}[t]
    \centering
    \includegraphics[width=0.8\textwidth]{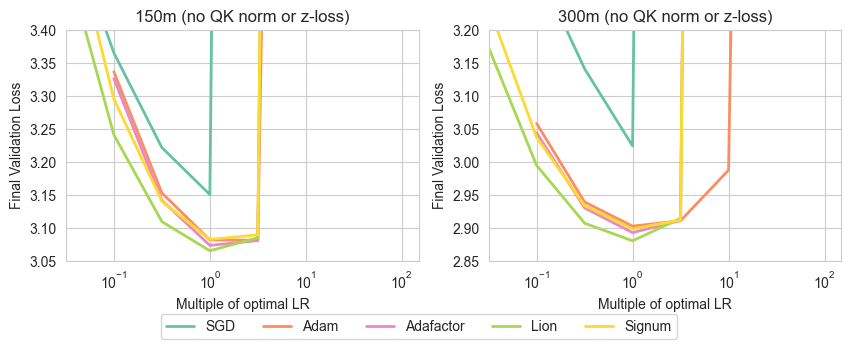}
    \caption{Sweeping learning rate without QK norm or z-loss for (\textbf{Left}) the 150m model, and (\textbf{Right}) the 300m model. These models are less stable than the standard model, but the same general trend across algorithms hold here.}
    \label{fig:lr_noqk}
\end{figure}

\begin{figure}[t]
    \centering
    \includegraphics[width=0.8\textwidth]{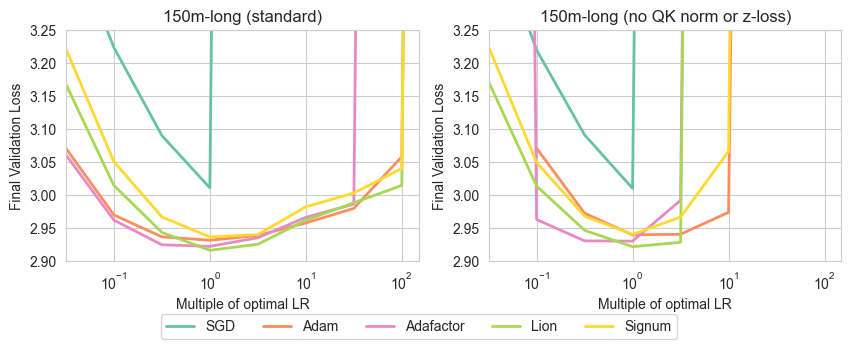}
    \caption{Sweeping learning rate on 150m models trained for 4x longer (100k steps) than in the base runs for (\textbf{Left}) the standard model, and (\textbf{Right}) the model without QK norm or z-loss. Compared to the shorter runs, these models achieve better performance and increased stability across learning rates.}
    \label{fig:lr_long}
\end{figure}

First, we sweep over the most important hyperparameter: learning rate. Note, in all of these sweeps over learning rate we set $ \beta_1 = 0.9$ for all algorithms except for SGD, where we set $ \beta_1 = 0.98$. As we will see in the following subsection, SGD is more sensitive to the momentum hyperparameters and requires more momentum to be competitive with the other optimizers.

Main results for our standard architecture across four scales are presented in \Cref{fig:main}. Note that the x-axis shifts the learning rates to align the optimal learning rates across algorithms. In terms of absolute learning rates, we sweep in multiples of $ \sqrt{10}$ from 1e-4 to 1 for Adam and Adafactor, from 1e-5 to 1e-1 for Lion and Signum, and from 1e-3 to 10 for SGD. We report on the optimal learning rates for each optimizer in \Cref{app:details}. 

The key takeaway is that not only do the algorithms achieve similar performance at the optimal learning rate, but the learning rate stability itself is similar across algorithms and scales. The one exception is SGD, which is worse both in terms of optimal performance and in terms of stability. 

Further ablations are presented in \Cref{fig:lr_noqk} and \Cref{fig:lr_long} illustrating performance for models with no QK norm or z-loss and 4x longer training time respectively. While we find that the architecture choices can clearly impact the amount of stability to learning rate, the cross-algorithm comparisons remain the same: Adafactor and Lion are competitive with Adam, while SGD is worse both in terms of performance and stability to learning rate. Similarly, training for longer can improve performance and stability to learning rate, but does not change the high-level cross-algorithm comparisons.

\paragraph{Takeaway:} performance and stability to learning rate are comparable across the non-SGD algorithms that we tested. 

\subsection{Sweeping momentum}

\begin{figure}[t]
    \centering
    \includegraphics[width=\textwidth]{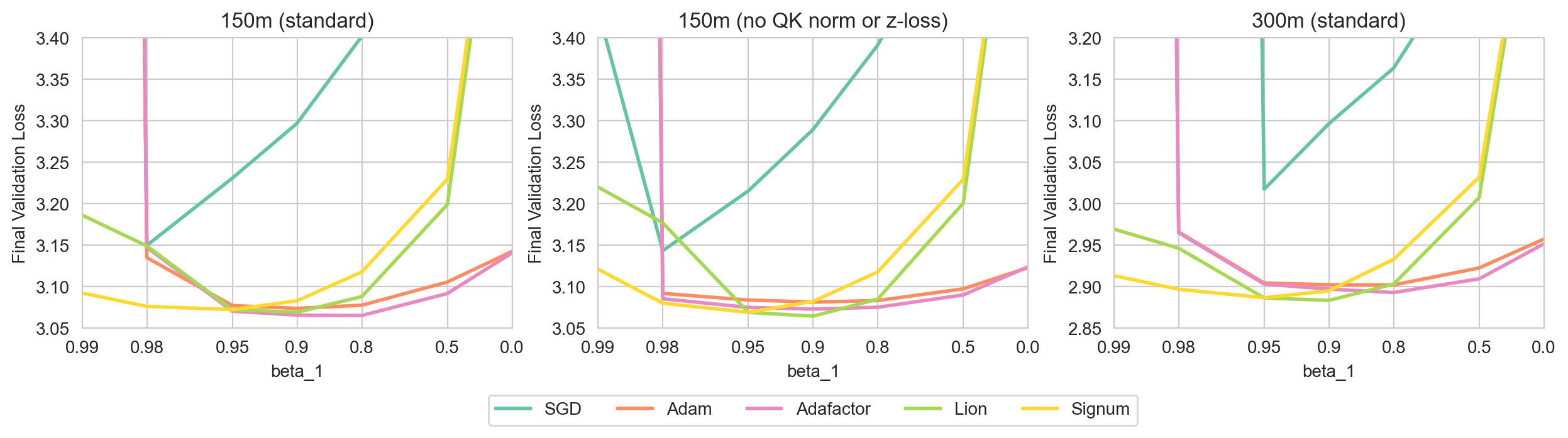}
    \caption{Sweeping momentum for fixed learning rate across three settings: (\textbf{Left}) 150m standard, (\textbf{Middle}) 150m with no QK norm or z-loss, (\textbf{Right}) 300m standard. Adam and Adafactor are similarly robust to $ \beta_1$, while Lion and Signum are slightly more sensitive to low values and SGD is substantially more sensitive.}
    \label{fig:momentum}
\end{figure}

Now we also sweep across momentum values (i.e. $ \beta_1$)\footnote{Note that in Lion, both $ \beta_1$ and $ \beta_2$ can be thought of as different types of ``momentum'' with $ \beta_1$ being the ``one-step'' momentum and $ \beta_2$ the ``long-term'' momentum. For consistency, we only sweep $ \beta_1$ here and sweep $\beta_2$ in Section~\ref{app:ablations}.}. To do this sweep we fix the per-algorithm learning rate to be the optimal learning rate from the corresponding learning rate sweep. 

Results are presented in \Cref{fig:momentum}. We observe that across various settings, the robustness to $ \beta_1$ is similar across the non-SGD algorithms when we stay in the range of momentums between 0.8 and 0.98. However, for high $ \beta_1$ Lion is better and low $ \beta_1$ Adam and Adafactor are better. Again we observe SGD being very sensitive to momentum.

\paragraph{Takeaway:} performance and stability to momentum are comparable across the non-SGD algorithms that we tested if we stay within the usual range of momentum values.

\subsection{Additional hyperparameter sweeps}\label{app:ablations}

\begin{figure}[h]
    \centering
    \includegraphics[width=\textwidth]{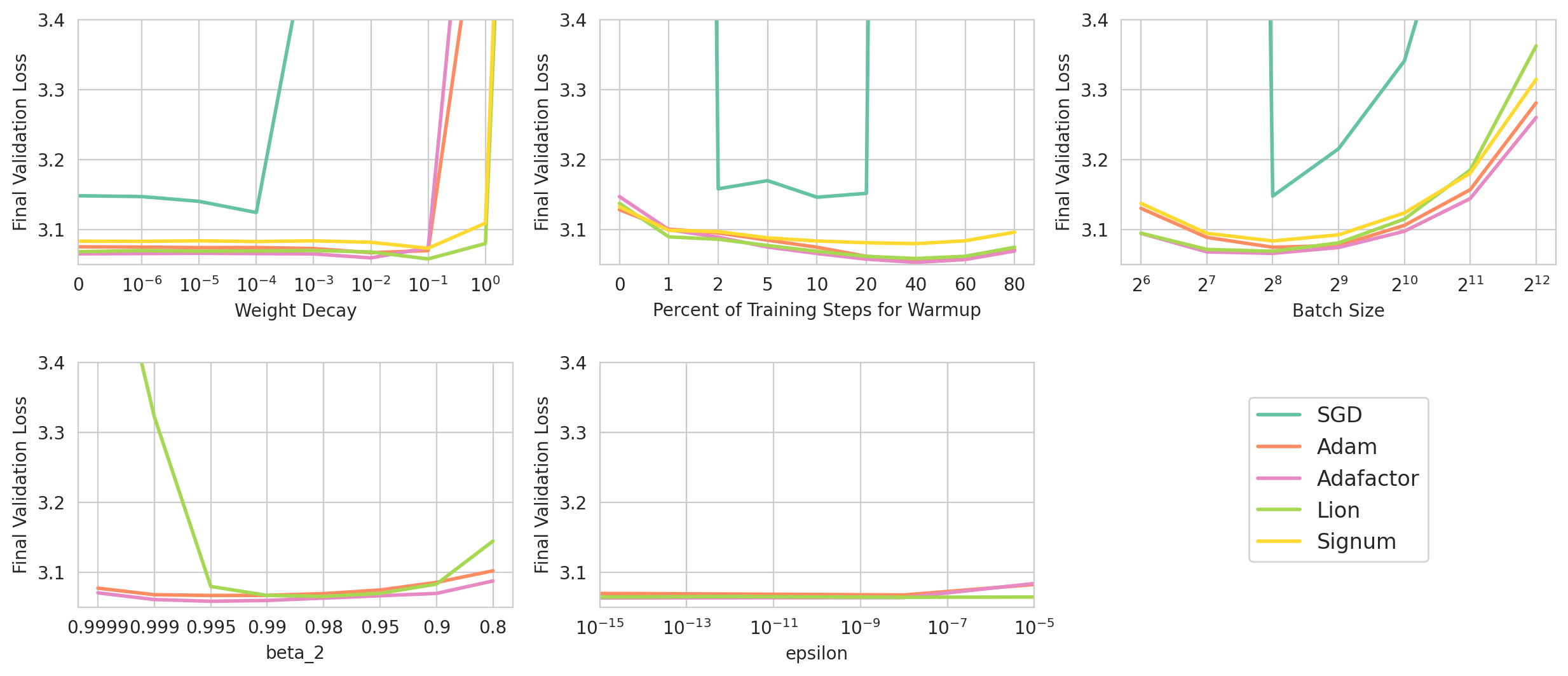}
    \caption{Sweeps over other hyperparameters. Top: weight decay, warmup duration, and batch size. Bottom: $\epsilon$ and $\beta_2$. We generally find little effect for the non-SGD algorithms, however there are parameters that differ from our defaults that can offer up to 0.02 improvements in validation loss}.
    \label{fig:wd}
\end{figure}

We also sweep over a variety of other hyperparameters in Figure \ref{fig:wd} using the best per-algorithm learning rate and momentum. We observe that SGD is less stable with respect to weight decay and warmup length. And while it is possible to get small benefits from higher weight decay, longer warmup, and higher $ \beta_2$ than our defaults, generally the algorithms are much more stable to these parameters than learning rate and momentum. We observe two exceptions: Lion shows poorer performance at extreme values of $\beta_2$. However, it is important to note that the $\beta_2$ parameter in Lion functions more like a form of ``momentum", whereas in Adam and Adafactor, $\beta_2$ regulates the moving average of the squared gradients. We also see that the performance across optimizers worsens at high batch size. This is expected as optimization algorithms are known to exhibit diminishing performance with increasing batch size \citep{shallue19, zhang2024doescriticalbatchsize}.

\paragraph{Takeaway:} generally algorithms are more stable with respect to other hyperparameters--- with the exception of batch size--- and the possible gains in performance are relatively small compared to learning rate and momentum.

\subsection{Signum recovers the performance and stability of Adam} \label{sec:signsgd}

\begin{figure}[h]
    \centering
    \includegraphics[width=\textwidth]{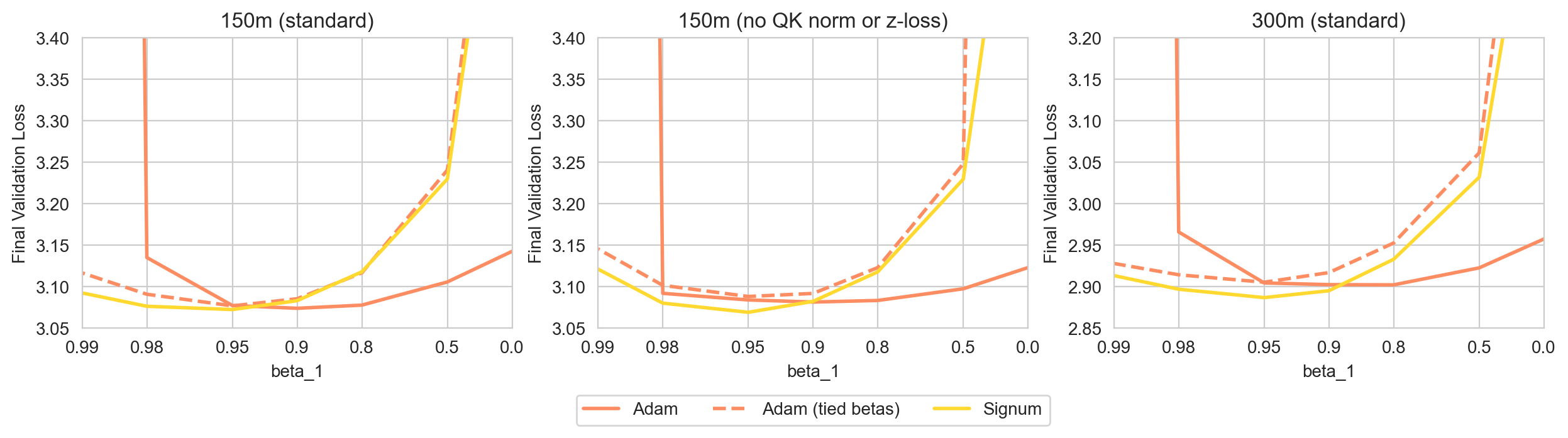}
    \caption{Sweeping momentum with $ \beta_1 = \beta_2$ tied together for Adam (dashed) and compared to Signum and Adam with fixed $ \beta_2 = 0.95$ (solid) across three settings: (\textbf{Left}) 150m standard, (\textbf{Middle}) 150m with no QK norm or z-loss, (\textbf{Right}) 300m standard. When $ \beta_1 = \beta_2$, Adam behaves very similarly to Signum. }
    \label{fig:eq_momentum}
\end{figure}

In Figure~\ref{fig:main} we observed that Adam and Signum have similar performance and stability for language modeling, even at scale. The following lemma from prior work\citep{balles2018dissecting} shows that Adam performs variance-adjusted sign gradient descent.

\begin{lemma}[~\cite{balles2018dissecting}]
\label{lem:beta}
    Consider a parameter with a history of gradients $g_t, g_{t-1}, \ldots$. Let $m$ be the random variable that is equal to $g_{t-\tau}$ with probability $(1-\beta_1)\beta_1^{\tau}$ and $v$ be the random variable that is equal to $g_{t-\tau}$ with probability $(1-\beta_2)\beta_2^{\tau}$. The Adam update $\delta_{\text{Adam}}$ and the Signum update $\delta_{\text{Signum}}$ are related by $$ \delta_{\text{Adam}} = \delta_{\text{Signum}} \cdot \frac{|\E[m]|}{\sqrt{\E[v^2]}}$$

\end{lemma}

If $\beta_1 = \beta_2$ then $m = v$ in Lemma~\ref{lem:beta} and hence the parameter-wise ratio of Adam and Signum updates is equal to the ratio of the mean and the square root of second moment of $m$. Intuitively, this holds because when $\beta_1 = \beta_2$, the first moment estimates of Signum and Adam, and second moment estimates of Adam, average the previous gradients with same coefficients ($(1-\beta)\beta^\tau$). This intuitively suggests that when $\beta_1 = \beta_2$, Adam and Signum may behave similarly \footnote{This is not true in theory, as the ratio can vary across parameters.}. This motivates the conjecture that the main benefit of Adam over Signum is the fact that in Adam, $\beta_2$ can be varied independently of $\beta_1$. In Figure~\ref{fig:main} we have $\beta_2 = 0.95$ and $\beta_1 = 0.9$ which are close, and as pointed out earlier, both optimizers have similar performance and stability. 

We examine this hypothesis further in Figure~\ref{fig:eq_momentum} by varying $\beta_1$ and setting $\beta_2 = \beta_1$, and again find that Signum and Adam behave very similarly. However, we also note that when we vary $\beta_1$ for Adam while fixing $\beta_2$ we get more stability for $\beta_1$ as compared to Signum.

\textbf{Takeaway: } With $\beta_2 = \beta_1$ Adam and Signum behave similarly and the standard setting for training language models ($\beta_2 = 0.95, \beta_1 = 0.9$) is close to this.

\section{Investigating the impact of adaptivity for optimizer stability and performance}
\label{sec:novograd_investigation}

Ablations in the previous section revealed the striking similarity in performance and stability across multiple optimizers compared to Adam. Adam and its other variants are designed to have a high degree of adaptivity at a fine granularity (per-parameter learning rates) throughout the training process. This adaptivity is often credited with the stability and robust performance observed in these optimizers. However, a critical question arises: to what extent is this adaptivity needed for different parameters of the network? By identifying the necessity of adaptivity for different network components to ensure both performance and stability, we aim to discern whether simpler optimizers like SGD can achieve similar benefits with minimal modifications. Since higher momentum can often play the same role as a better preconditioner and to have all algorithms on an equal footing, we will fix $\beta_1 = 0.9$ for all optimizers in this section.

The main optimizer we study in this section is a ``layer-wise'' variant of Adam, which we coin as `Adalayer'.  We use Adalayer for our investigations because it lends a greater ease of understanding compared to full-fledged Adam in identifying parts of the network which may be particularly critical for optimizer performance and stability. Note that this layerwise variant is a special case of a previously known optimizer called Blockwise Adaptive Gradient with Momentum (BAGM) \citep{bagm}.

\subsection{Adalayer}
\label{subsec:lastlayer_correction}

\begin{wrapfigure}[13]{r}{0.55\textwidth}
\begin{minipage}{0.55\textwidth}
\vspace{-1.4cm}
\begin{algorithm}[H]
\SetAlgoNlRelativeSize{-1} %
\SetNlSty{textbf}{\{}{\}} %
\caption{Adalayer \label{alg:adalayer}}
\textbf{Parameters:}
Learning rate $\eta$, exponential decay rates for the moment estimates $\beta_1, \beta_2$, number of steps $T$, $\epsilon$

\While{$t \leq T$}{
    \For{each layer $l$ with $p$ parameters}{
        $g_t^l \gets \nabla_l L(w_t)$ \;
        
        $v_t^l \gets \beta_2 \cdot v_{t-1}^l + (1 - \beta_2) \cdot p^{-1/2} \cdot \|g_t^l\|_2^2$ \;
        
        $m_t^l \gets \beta_1 \cdot m_{t-1}^l + (1-\beta_1)g_t^l$ \;
        
        $w_{t+1}^l \gets w_t^l - \eta \cdot \frac{m_t^l}{\sqrt{v_t^l}+\epsilon}$ \;
    }
}
\end{algorithm}
\end{minipage}
\end{wrapfigure}

To study the behavior of adaptive optimizers like Adam, we begin with describing a layer-wise version of Adam which we refer to as Adalayer. Adam, Adafactor and Adalayer all (approximately) store the diagonal second moment matrix, but with coarser and coarser granularity; for a layer of dimension $m \times n$, Adam explicitly maintains the second moment matrix using $mn$ parameters in the shape of a matrix. Adafactor stores row and column averages of the second moment matrix which serve as a rank-1 approximation to the second moment matrix. Finally, Adalayer stores a single scalar which is the average of the second moment matrix. We will later consider a generalization of Adalayer where instead of averaging second moment over a layer we will average it over a ``block'' of parameters which can be a subset of a layer. We note that similar algorithms have been studied before ~\citep{ginsburg2019stochastic,grafting} but we choose to study this variant since it is a direct analogue of Adam and Adafactor. A simplified version of Adalayer optimizer is given in Algorithm~\ref{alg:adalayer}; other details such as bias correction are kept same as that for Adam. In \Cref{app:adalayer}, we provide more details about our Adalayer implementation and a correction we make to how Adalayer treats the last layer in order to achieve similar performance and stability to Adam. Specifically, Adalayer when naively applied for each layer is neither performant nor stable to learning rate (\Cref{fig:novograd_last_correction}); however, if we additionally treat the set of weights in the last layer feeding into each logit as its own block, this recovers most of the performance and stability of Adam (see the dotted blue lines in Figure~\ref{fig:sgd+novograd}). We henceforth refer to Adalayer with this correction as \textbf{Adalayer*}. %

\begin{figure}[t]
    \centering
    \includegraphics[width=\textwidth]{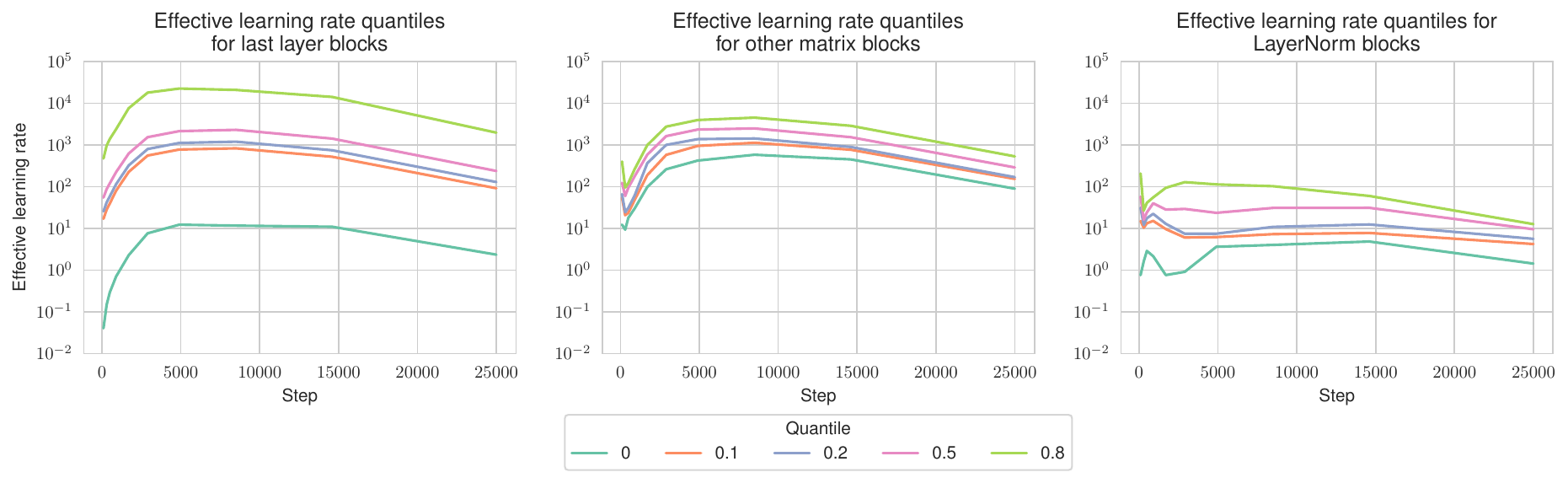}
    \caption{Quantiles of effective learning rates ($\eta_t / (\sqrt{v^l_t}+\epsilon)$ for each layer $l$) for the last layer blocks (\textbf{Left}), the LayerNorm blocks (\textbf{Right}), and the other matrix blocks (\textbf{Middle}) for a 150m model trained using Adalayer*. Unlike the other matrix blocks and LayerNorm parameters, the effective learning rates across logits vary across multiple orders of magnitude, providing evidence for the need to precondition them separately.}
    \label{fig:eff-lr-all}
\end{figure}

To study how Adalayer* preconditions the network, we plot effective learning rates used for different logits by Adalayer* in Figure~\ref{fig:eff-lr-all} (\textbf{Left}). Here, the effective learning rate for a layer $l$ in the network is $\eta_t / (\sqrt{v^l_t}+\epsilon)$. We find that Adalayer* indeed uses vastly different learning rates for different logits, supporting our hypothesis that preconditioning weight in different logits separately is important for performance and stability.

\subsection{Both the last layer and LayerNorm parameters need adaptivity}
\label{subsec:sgd+novograd}
The results using Adalayer* in the previous section suggest that all layers except the last layer only need a iteration-dependent scalar correction to their learning rate. We now ask a stronger question: do we need these scales at all? Or can we train the remaining layers with SGD? This hypothesis is supported by looking at Figure~\ref{fig:eff-lr-all} (middle) where we observe that the learning rates for different matrix layers (except the last layer) assigned by Adalayer* are remarkably similar.

\begin{figure}[h]
    \centering
    \includegraphics[width=0.48\textwidth]{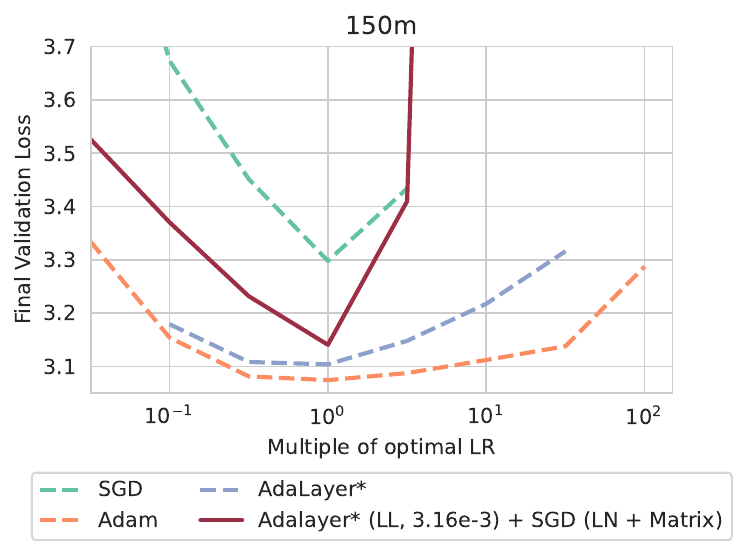}
    \includegraphics[width=0.48\textwidth]{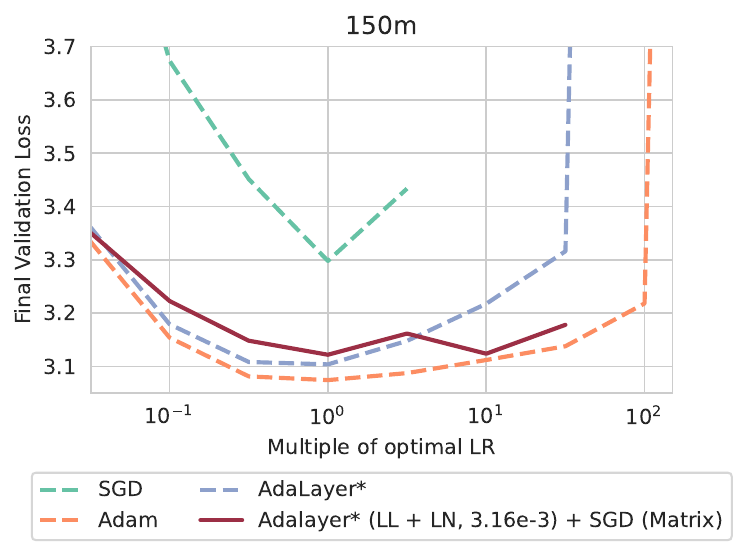}
    \caption{(\textbf{Left}): Training the last layer using Adalayer* with a fixed learning rate of $3.16e-3$ and other LayerNorm and matrix blocks using SGD achieves better performance than SGD, but does not recover stability. (\textbf{Right}): Training both the last layer and LayerNorm blocks using Adalayer* and the other matrix blocks using SGD nearly recovers or exceeds performance of Adalayer*, and achieves stability across learning rates. Dotted lines are baselines from optimizers previously given in Sections \ref{sec:ablations} and \ref{subsec:lastlayer_correction}.}
    \label{fig:sgd+novograd}
\end{figure}

To test this, we train the last layer with Adalayer* (fixing a learning rate of $3.16e-3$) and the rest of the layers with SGD, both with $\beta_1 = 0.9$. In Figure~\ref{fig:sgd+novograd} (\textbf{Left}) we show the results while sweeping over SGD learning rates from $0.1$ to $3160$. While this improves upon the performance of SGD, we do not recover stability of the Adalayer* and Adam baselines. We trace this instability to \emph{LayerNorm blocks}: Figure~\ref{fig:eff-lr-all} (\textbf{Left}) shows that the effective learning rates for the LayerNorm blocks are much smaller, which suggests that they may destabilize at higher SGD learning rates. To ameliorate this, in Figure~\ref{fig:sgd+novograd} (\textbf{Right}) we add LayerNorm parameters to those being trained with Adalayer* and find that this is sufficient to recover both performance and stability of Adalayer*. In Figure~\ref{fig:sgd_novograd_600} in the Appendix, we see this trend continue to hold for 300m and 600m parameter models. 

We conduct additional experiments investigating these `hybrid' variants of SGD in the Appendix. Despite most language model parameters being in matrix layers, we show that if we only apply Adalayer* on the matrix layers and train the last layer and LayerNorm layers with SGD, this does not recover performance and stability. Another plausible remedy to address the small effective learning rates of the LayerNorm blocks is to simply not train these LayerNorm parameters. In Appendix~\ref{app:sgd_variants}, Figure~\ref{fig:sgd_novograd_ln_off}, we show that this improves performance and stability relative to SGD but does not recover Adalayer* performance. Finally in Appendix~\ref{app:adasgd}, we replace SGD with AdaSGD~\citep{wang2020adasgd} which still uses a global learning rate but is now adaptive through training; in Figure~\ref{fig:adasgd_adalayer} we show AdaSGD + Adalayer* can recover full Adalayer* performance, but a small gap remains to recover Adam performance.

We acknowledge that while adaptivity for the last layer and LayerNorm parameters seems \textit{necessary} to retain optimizer performance and stability, there remains a small gap which is fulfilled by adaptivity on the other network parameters; in Figure~\ref{fig:sgd_adafactor} we also try training 150m and 300m models using SGD on the matrix blocks and \textit{Adafactor} on the last layer and LayerNorm blocks; we find that performance and stability is comparable, but does not exceed that of Adafactor on the entire network.

Further, note that a caveat of the above results is that we have introduced an additional hyperparameter--- SGD learning rate, which we are sweeping over--- while keeping the Adalayer* learning rate in the last layer and layer norm layers fixed. While decoupling the learning rates here is needed (due to SGD's performant learning rates being orders of magnitude higher than that of Adalayer*), this may be responsible for the observed stability. To address this, we perform the following experiment: we train all the layers with Adalayer* (sweeping over learning rate $\eta$) but we \emph{stop updating the second moment estimates} for all layers \emph{except the last layer and LayerNorm blocks} after initialization. This implies that these layers are effectively being trained by SGD with a fixed learning rate, though unlike the above results, these learning rates are different for different layers. We implement this by passing 1000 batches to initialized 150m and 300m models to obtain second moment estimates for all layers without letting the model take a gradient step, and then allowing the model to train as normal under the same settings as all of our ablations. As in our previous investigation, we fix other hyperparameters to be the same: $\beta_1 = 0.9$, $\beta_2 = 0.95$ and $\epsilon = 1e-15$.

\begin{figure}[t]
    \centering
    \includegraphics[width=0.48\textwidth]{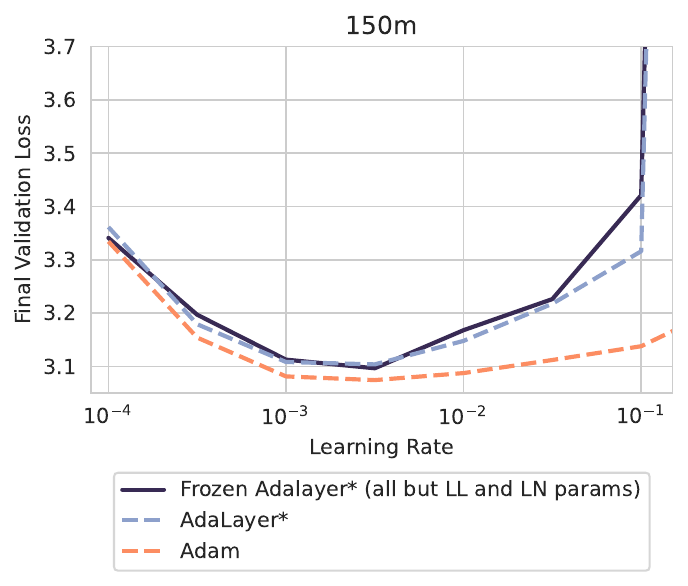}
    \includegraphics[width=0.48\textwidth]{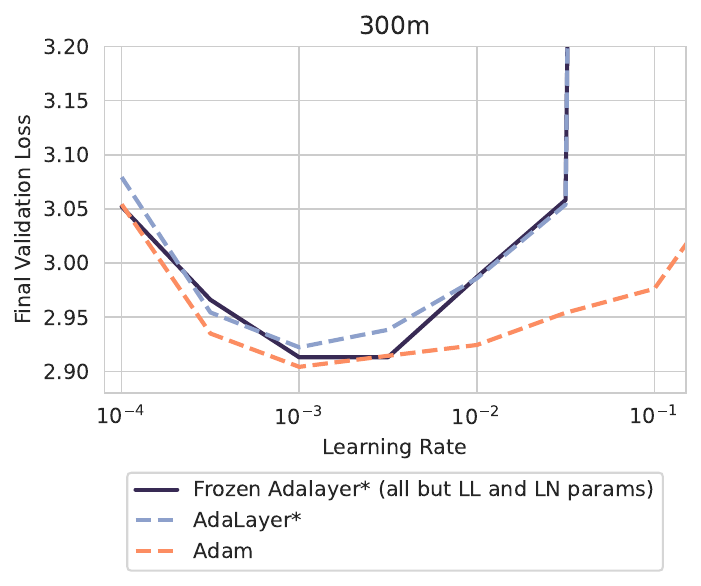}
    \caption{Training 150m (\textbf{Left}) and 300m (\textbf{Right}) models using fixed Adalayer* learning rate ratios from initialization, with the exception of last layer and LayerNorm parameters. This almost entirely matches the performance and stability of Adalayer* in the 150m model, and exceeds Adalayer*'s peak performance to be comparable with Adam.}
    \label{fig:novograd_freeze}
\end{figure}

In Figure~\ref{fig:novograd_freeze} we show the resulting learning rate sweep for freezing Adalayer* learning rate scales at initialization (with the exception of the last layer and LayerNorm). Surprisingly, we find for the 150m model that we can almost entirely recover the stability and performance of Adalayer*. For the 300m model we also match or exceed the performance of Adalayer*, and even nearly match the peak performance of Adam. Note again that this sweeps learning rate across all network parameters. This provides further evidence for the importance of adaptivity in the last layer and LayerNorm, where in contrast we could have used fixed ratios from initialization for all other parameters to recover the performance and stability of Adalayer*. In Appendix~\ref{app:novograd_freeze} we present additional experiments which include allowing Adalayer* to update for only one of the last layer or LayerNorm, or turning off LayerNorm training. None of these partial modifications fully recover performance and stability.

\section{Discussion and Limitations}
\label{sec:discussion}

After a comprehensive comparison of a variety of optimizers for language modeling, we have found that many optimizers seem to be roughly equivalent both in terms of optimal performance and hyperparameter stability. The only differing axis for practitioners to consider is thus memory constraints, but based on our results, memory considerations for optimizer choice should not affect the performance of their training run, at least across the optimizers tested (i.e. optimizers using diagonal preconditioning). For tuning guidelines, we have reported the optimal learning rates in Appendix~\ref{app:details} found in our sweeps as a starting point for practitioners in tuning. We also observed that most other parameters are very stable for the optimal learning rate (with exception to momentum) and thus we believe tuning learning rate and momentum should be prioritized if one has a limited compute budget for hyperparameter sweeping. 

Diving deeper, we have shown that the treatment of the last layer and LayerNorm parameters is crucial for realizing the benefits of adaptive optimizers. This highlights the following research directions as the most promising for developing optimizers for language model training. Firstly, our results suggest that within the realm of diagonal preconditioning optimizers, there may be room for more memory-efficient alternatives to AdamW to be designed beyond using factorization techniques (eg. Adafactor with momentum and Lion from our initial sweeps). We have provided one direction for cutting down on memory overhead, where we have shown that the need for adaptivity is distributed unevenly across different parameters of the network (with similar ideas already explored in concurrent work~\citep{zhang2024adamminiusefewerlearning}). Secondly, in terms of improving the performance of these models, our work suggests that optimizer choice is unlikely to be a promising approach--- if we are limited to diagonal preconditioning optimizers. Thus, it is likely that we need to turn to non-diagonal methods to improve on our current methods; this is indeed where previous work has demonstrated improved performance to Adam~\citep{gupta2018shampoo}.

Of course, there are several limitations to our study including the fact that due to computational constraints we only ablate a few architecture decisions, that we only consider one dimensional hyperparameter sweeps, we fix batch size, and that we limit our study to autoregressive language modeling with a single dataset. Despite these limitations, we believe that the study sheds new light on the fundamentals of optimization for language modeling.

\newpage
\section*{Acknowledgments}

SK, DM and RZ acknowledges support from the Office of Naval Research under award N00014-22-1-2377 and the National Science Foundation Grant under award \#IIS 2229881. This work has been made possible in part by a gift from the Chan Zuckerberg Initiative Foundation to establish the Kempner Institute for the Study of Natural and Artificial Intelligence. NV, DM and RZ are supported by a Simons Investigator Fellowship, NSF grant DMS-2134157, DARPA grant W911NF2010021,and DOE grant DE-SC0022199. RZ and DM are supported by Kempner Institute Graduate Research Fellowships.

\bibliography{references}

\begin{thebibliography}{47}
\providecommand{\natexlab}[1]{#1}
\providecommand{\url}[1]{\texttt{#1}}
\expandafter\ifx\csname urlstyle\endcsname\relax
  \providecommand{\doi}[1]{doi: #1}\else
  \providecommand{\doi}{doi: \begingroup \urlstyle{rm}\Url}\fi

\bibitem[Agarwal et~al.(2020)Agarwal, Anil, Hazan, Koren, and Zhang]{grafting}
Naman Agarwal, Rohan Anil, Elad Hazan, Tomer Koren, and Cyril Zhang.
\newblock Disentangling adaptive gradient methods from learning rates.
\newblock \emph{CoRR}, abs/2002.11803, 2020.
\newblock URL \url{https://arxiv.org/abs/2002.11803}.

\bibitem[Ahn et~al.(2024)Ahn, Cheng, Song, Yun, Jadbabaie, and Sra]{ahn2024linear}
Kwangjun Ahn, Xiang Cheng, Minhak Song, Chulhee Yun, Ali Jadbabaie, and Suvrit Sra.
\newblock Linear attention is (maybe) all you need (to understand transformer optimization).
\newblock In \emph{The Twelfth International Conference on Learning Representations}, 2024.
\newblock URL \url{https://openreview.net/forum?id=0uI5415ry7}.

\bibitem[Balles \& Hennig(2018)Balles and Hennig]{balles2018dissecting}
Lukas Balles and Philipp Hennig.
\newblock Dissecting adam: The sign, magnitude and variance of stochastic gradients, 2018.
\newblock URL \url{https://openreview.net/forum?id=S1EwLkW0W}.

\bibitem[Balles et~al.(2020)Balles, Pedregosa, and Roux]{balles2020geometry}
Lukas Balles, Fabian Pedregosa, and Nicolas~Le Roux.
\newblock The geometry of sign gradient descent, 2020.

\bibitem[Bartz-Beielstein et~al.(2020)Bartz-Beielstein, Doerr, Berg, Bossek, Chandrasekaran, Eftimov, Fischbach, Kerschke, La~Cava, Lopez-Ibanez, et~al.]{bartz2020benchmarking}
Thomas Bartz-Beielstein, Carola Doerr, Daan van~den Berg, Jakob Bossek, Sowmya Chandrasekaran, Tome Eftimov, Andreas Fischbach, Pascal Kerschke, William La~Cava, Manuel Lopez-Ibanez, et~al.
\newblock Benchmarking in optimization: Best practice and open issues.
\newblock \emph{arXiv preprint arXiv:2007.03488}, 2020.

\bibitem[Bernstein et~al.(2018)Bernstein, Wang, Azizzadenesheli, and Anandkumar]{pmlr-v80-bernstein18a}
Jeremy Bernstein, Yu-Xiang Wang, Kamyar Azizzadenesheli, and Animashree Anandkumar.
\newblock sign{SGD}: Compressed optimisation for non-convex problems.
\newblock In Jennifer Dy and Andreas Krause (eds.), \emph{Proceedings of the 35th International Conference on Machine Learning}, volume~80 of \emph{Proceedings of Machine Learning Research}, pp.\  560--569. PMLR, 10--15 Jul 2018.
\newblock URL \url{https://proceedings.mlr.press/v80/bernstein18a.html}.

\bibitem[Bernstein et~al.(2019)Bernstein, Zhao, Azizzadenesheli, and Anandkumar]{bernstein2018signsgd}
Jeremy Bernstein, Jiawei Zhao, Kamyar Azizzadenesheli, and Anima Anandkumar.
\newblock sign{SGD} with majority vote is communication efficient and fault tolerant.
\newblock In \emph{International Conference on Learning Representations}, 2019.
\newblock URL \url{https://openreview.net/forum?id=BJxhijAcY7}.

\bibitem[Chen et~al.(2023)Chen, Liang, Huang, Real, Wang, Pham, Dong, Luong, Hsieh, Lu, and Le]{chen2023symbolic}
Xiangning Chen, Chen Liang, Da~Huang, Esteban Real, Kaiyuan Wang, Hieu Pham, Xuanyi Dong, Thang Luong, Cho-Jui Hsieh, Yifeng Lu, and Quoc~V Le.
\newblock Symbolic discovery of optimization algorithms.
\newblock In \emph{Thirty-seventh Conference on Neural Information Processing Systems}, 2023.
\newblock URL \url{https://openreview.net/forum?id=ne6zeqLFCZ}.

\bibitem[Chowdhery et~al.(2023)Chowdhery, Narang, Devlin, Bosma, Mishra, Roberts, Barham, Chung, Sutton, Gehrmann, et~al.]{chowdhery2023palm}
Aakanksha Chowdhery, Sharan Narang, Jacob Devlin, Maarten Bosma, Gaurav Mishra, Adam Roberts, Paul Barham, Hyung~Won Chung, Charles Sutton, Sebastian Gehrmann, et~al.
\newblock Palm: Scaling language modeling with pathways.
\newblock \emph{Journal of Machine Learning Research}, 24\penalty0 (240):\penalty0 1--113, 2023.

\bibitem[Dehghani et~al.(2023)Dehghani, Djolonga, Mustafa, Padlewski, Heek, Gilmer, Steiner, Caron, Geirhos, Alabdulmohsin, et~al.]{dehghani2023scaling}
Mostafa Dehghani, Josip Djolonga, Basil Mustafa, Piotr Padlewski, Jonathan Heek, Justin Gilmer, Andreas~Peter Steiner, Mathilde Caron, Robert Geirhos, Ibrahim Alabdulmohsin, et~al.
\newblock Scaling vision transformers to 22 billion parameters.
\newblock In \emph{International Conference on Machine Learning}, pp.\  7480--7512. PMLR, 2023.

\bibitem[Ginsburg et~al.(2019)Ginsburg, Castonguay, Hrinchuk, Kuchaiev, Lavrukhin, Leary, Li, Nguyen, Zhang, and Cohen]{ginsburg2019stochastic}
Boris Ginsburg, Patrice Castonguay, Oleksii Hrinchuk, Oleksii Kuchaiev, Vitaly Lavrukhin, Ryan Leary, Jason Li, Huyen Nguyen, Yang Zhang, and Jonathan~M Cohen.
\newblock Stochastic gradient methods with layer-wise adaptive moments for training of deep networks.
\newblock \emph{arXiv preprint arXiv:1905.11286}, 2019.

\bibitem[Groeneveld et~al.(2024)Groeneveld, Beltagy, Walsh, Bhagia, Kinney, Tafjord, Jha, Ivison, Magnusson, Wang, et~al.]{groeneveld2024olmo}
Dirk Groeneveld, Iz~Beltagy, Pete Walsh, Akshita Bhagia, Rodney Kinney, Oyvind Tafjord, Ananya~Harsh Jha, Hamish Ivison, Ian Magnusson, Yizhong Wang, et~al.
\newblock Olmo: Accelerating the science of language models.
\newblock \emph{arXiv preprint arXiv:2402.00838}, 2024.

\bibitem[Gupta et~al.(2018{\natexlab{a}})Gupta, Koren, and Singer]{gupta2018shampoo}
Vineet Gupta, Tomer Koren, and Yoram Singer.
\newblock Shampoo: Preconditioned stochastic tensor optimization.
\newblock In \emph{International Conference on Machine Learning}, pp.\  1842--1850. PMLR, 2018{\natexlab{a}}.

\bibitem[Gupta et~al.(2018{\natexlab{b}})Gupta, Koren, and Singer]{shampoo}
Vineet Gupta, Tomer Koren, and Yoram Singer.
\newblock Shampoo: Preconditioned stochastic tensor optimization.
\newblock In Jennifer~G. Dy and Andreas Krause (eds.), \emph{Proceedings of the 35th International Conference on Machine Learning, {ICML} 2018, Stockholmsm{\"{a}}ssan, Stockholm, Sweden, July 10-15, 2018}, volume~80 of \emph{Proceedings of Machine Learning Research}, pp.\  1837--1845. {PMLR}, 2018{\natexlab{b}}.
\newblock URL \url{http://proceedings.mlr.press/v80/gupta18a.html}.

\bibitem[Hendrycks \& Gimpel(2016)Hendrycks and Gimpel]{hendrycks2016gaussian}
Dan Hendrycks and Kevin Gimpel.
\newblock Gaussian error linear units (gelus).
\newblock \emph{arXiv preprint arXiv:1606.08415}, 2016.

\bibitem[Jelassi et~al.(2022)Jelassi, Dobre, Mensch, Li, and Gidel]{jelassi2022dissecting}
Samy Jelassi, David Dobre, Arthur Mensch, Yuanzhi Li, and Gauthier Gidel.
\newblock Dissecting adaptive methods in gans, 2022.

\bibitem[Jiang et~al.(2023)Jiang, Malik, and Li]{jiang2023how}
Kaiqi Jiang, Dhruv Malik, and Yuanzhi Li.
\newblock How does adaptive optimization impact local neural network geometry?
\newblock In \emph{Thirty-seventh Conference on Neural Information Processing Systems}, 2023.
\newblock URL \url{https://openreview.net/forum?id=gIG8LvTLuc}.

\bibitem[Kaddour et~al.(2024)Kaddour, Key, Nawrot, Minervini, and Kusner]{kaddour2024no}
Jean Kaddour, Oscar Key, Piotr Nawrot, Pasquale Minervini, and Matt~J Kusner.
\newblock No train no gain: Revisiting efficient training algorithms for transformer-based language models.
\newblock \emph{Advances in Neural Information Processing Systems}, 36, 2024.

\bibitem[Kingma \& Ba(2015)Kingma and Ba]{kingma15}
Diederik~P. Kingma and Jimmy Ba.
\newblock Adam: {A} method for stochastic optimization.
\newblock In Yoshua Bengio and Yann LeCun (eds.), \emph{3rd International Conference on Learning Representations, {ICLR} 2015, San Diego, CA, USA, May 7-9, 2015, Conference Track Proceedings}, 2015.
\newblock URL \url{http://arxiv.org/abs/1412.6980}.

\bibitem[Kumar et~al.(2024)Kumar, Shen, Bubeck, and Gunasekar]{kumar2024how}
Ananya Kumar, Ruoqi Shen, Sebastien Bubeck, and Suriya Gunasekar.
\newblock How to fine-tune vision models with {SGD}.
\newblock In \emph{The Twelfth International Conference on Learning Representations}, 2024.
\newblock URL \url{https://openreview.net/forum?id=ZTssMmhC2X}.

\bibitem[Kunstner et~al.(2023)Kunstner, Chen, Lavington, and Schmidt]{kunstner2023noise}
Frederik Kunstner, Jacques Chen, Jonathan~Wilder Lavington, and Mark Schmidt.
\newblock Noise is not the main factor behind the gap between sgd and adam on transformers, but sign descent might be.
\newblock In \emph{The Eleventh International Conference on Learning Representations}, 2023.
\newblock URL \url{https://openreview.net/forum?id=a65YK0cqH8g}.

\bibitem[Kunstner et~al.(2024)Kunstner, Yadav, Milligan, Schmidt, and Bietti]{kunstner2024heavy}
Frederik Kunstner, Robin Yadav, Alan Milligan, Mark Schmidt, and Alberto Bietti.
\newblock Heavy-tailed class imbalance and why adam outperforms gradient descent on language models.
\newblock \emph{CoRR}, 2024.

\bibitem[Liu et~al.(2024)Liu, Li, Hall, Liang, and Ma]{liu2024sophia}
Hong Liu, Zhiyuan Li, David Leo~Wright Hall, Percy Liang, and Tengyu Ma.
\newblock Sophia: A scalable stochastic second-order optimizer for language model pre-training.
\newblock In \emph{The Twelfth International Conference on Learning Representations}, 2024.
\newblock URL \url{https://openreview.net/forum?id=3xHDeA8Noi}.

\bibitem[Liu et~al.(2021)Liu, Bernstein, Meister, and Yue]{pmlr-v139-liu21c}
Yang Liu, Jeremy Bernstein, Markus Meister, and Yisong Yue.
\newblock Learning by turning: Neural architecture aware optimisation.
\newblock In Marina Meila and Tong Zhang (eds.), \emph{Proceedings of the 38th International Conference on Machine Learning}, volume 139 of \emph{Proceedings of Machine Learning Research}, pp.\  6748--6758. PMLR, 18--24 Jul 2021.
\newblock URL \url{https://proceedings.mlr.press/v139/liu21c.html}.

\bibitem[Merity et~al.(2017)Merity, Xiong, Bradbury, and Socher]{merity2017pointer}
Stephen Merity, Caiming Xiong, James Bradbury, and Richard Socher.
\newblock Pointer sentinel mixture models.
\newblock In \emph{International Conference on Learning Representations}, 2017.
\newblock URL \url{https://openreview.net/forum?id=Byj72udxe}.

\bibitem[Nesterov(2004)]{nesterov2004introductory}
I.U.E. Nesterov.
\newblock \emph{Introductory Lectures on Convex Optimization: A Basic Course}.
\newblock Mathematics and its applications. Kluwer Academic Publishers, 2004.
\newblock ISBN 9780004501444.
\newblock URL \url{https://books.google.com/books?id=klv2PwAACAAJ}.

\bibitem[Pan \& Li(2022)Pan and Li]{pan2022toward}
Yan Pan and Yuanzhi Li.
\newblock Toward understanding why adam converges faster than {SGD} for transformers.
\newblock In \emph{OPT 2022: Optimization for Machine Learning (NeurIPS 2022 Workshop)}, 2022.
\newblock URL \url{https://openreview.net/forum?id=Sf1NlV2r6PO}.

\bibitem[Paszke et~al.(2019)Paszke, Gross, Massa, Lerer, Bradbury, Chanan, Killeen, Lin, Gimelshein, Antiga, et~al.]{paszke2019pytorch}
Adam Paszke, Sam Gross, Francisco Massa, Adam Lerer, James Bradbury, Gregory Chanan, Trevor Killeen, Zeming Lin, Natalia Gimelshein, Luca Antiga, et~al.
\newblock Pytorch: An imperative style, high-performance deep learning library.
\newblock \emph{Advances in neural information processing systems}, 32, 2019.

\bibitem[Porian et~al.(2024)Porian, Wortsman, Jitsev, Schmidt, and Carmon]{porian2024resolving}
Tomer Porian, Mitchell Wortsman, Jenia Jitsev, Ludwig Schmidt, and Yair Carmon.
\newblock Resolving discrepancies in compute-optimal scaling of language models.
\newblock \emph{arXiv preprint arXiv:2406.19146}, 2024.

\bibitem[Raffel et~al.(2020)Raffel, Shazeer, Roberts, Lee, Narang, Matena, Zhou, Li, and Liu]{raffel2020exploring}
Colin Raffel, Noam Shazeer, Adam Roberts, Katherine Lee, Sharan Narang, Michael Matena, Yanqi Zhou, Wei Li, and Peter~J Liu.
\newblock Exploring the limits of transfer learning with a unified text-to-text transformer.
\newblock \emph{Journal of machine learning research}, 21\penalty0 (140):\penalty0 1--67, 2020.

\bibitem[Rajpurkar et~al.(2016)Rajpurkar, Zhang, Lopyrev, and Liang]{rajpurkar2016squad}
Pranav Rajpurkar, Jian Zhang, Konstantin Lopyrev, and Percy Liang.
\newblock Squad: 100,000+ questions for machine comprehension of text, 2016.

\bibitem[Robbins \& Monro(1951)Robbins and Monro]{robbinsmonro51}
Herbert Robbins and Sutton Monro.
\newblock {A Stochastic Approximation Method}.
\newblock \emph{The Annals of Mathematical Statistics}, 22\penalty0 (3):\penalty0 400 -- 407, 1951.
\newblock \doi{10.1214/aoms/1177729586}.
\newblock URL \url{https://doi.org/10.1214/aoms/1177729586}.

\bibitem[Schmidt et~al.(2021)Schmidt, Schneider, and Hennig]{schmidt2021descending}
Robin~M Schmidt, Frank Schneider, and Philipp Hennig.
\newblock Descending through a crowded valley-benchmarking deep learning optimizers.
\newblock In \emph{International Conference on Machine Learning}, pp.\  9367--9376. PMLR, 2021.

\bibitem[Schneider et~al.(2019)Schneider, Balles, and Hennig]{schneiderdeepobs}
Frank Schneider, Lukas Balles, and Philipp Hennig.
\newblock Deepobs: A deep learning optimizer benchmark suite.
\newblock In \emph{International Conference on Learning Representations}, 2019.

\bibitem[Shallue et~al.(2019)Shallue, Lee, Antognini, Sohl-Dickstein, Frostig, and Dahl]{shallue19}
Christopher~J. Shallue, Jaehoon Lee, Joseph Antognini, Jascha Sohl-Dickstein, Roy Frostig, and George~E. Dahl.
\newblock Measuring the effects of data parallelism on neural network training.
\newblock \emph{Journal of Machine Learning Research}, 20\penalty0 (112):\penalty0 1--49, 2019.
\newblock URL \url{http://jmlr.org/papers/v20/18-789.html}.

\bibitem[Shazeer \& Stern(2018)Shazeer and Stern]{ShazeerS18}
Noam Shazeer and Mitchell Stern.
\newblock Adafactor: Adaptive learning rates with sublinear memory cost.
\newblock In Jennifer~G. Dy and Andreas Krause (eds.), \emph{Proceedings of the 35th International Conference on Machine Learning, {ICML} 2018, Stockholmsm{\"{a}}ssan, Stockholm, Sweden, July 10-15, 2018}, volume~80 of \emph{Proceedings of Machine Learning Research}, pp.\  4603--4611. {PMLR}, 2018.
\newblock URL \url{http://proceedings.mlr.press/v80/shazeer18a.html}.

\bibitem[Sivaprasad et~al.(2020)Sivaprasad, Mai, Vogels, Jaggi, and Fleuret]{sivaprasad2020optimizer}
Prabhu~Teja Sivaprasad, Florian Mai, Thijs Vogels, Martin Jaggi, and Fran{\c{c}}ois Fleuret.
\newblock Optimizer benchmarking needs to account for hyperparameter tuning.
\newblock In \emph{International conference on machine learning}, pp.\  9036--9045. PMLR, 2020.

\bibitem[Su et~al.(2024)Su, Ahmed, Lu, Pan, Bo, and Liu]{su2024roformer}
Jianlin Su, Murtadha Ahmed, Yu~Lu, Shengfeng Pan, Wen Bo, and Yunfeng Liu.
\newblock Roformer: Enhanced transformer with rotary position embedding.
\newblock \emph{Neurocomputing}, 568:\penalty0 127063, 2024.

\bibitem[Wang \& Wiens(2020)Wang and Wiens]{wang2020adasgd}
Jiaxuan Wang and Jenna Wiens.
\newblock Adasgd: Bridging the gap between sgd and adam.
\newblock \emph{arXiv preprint arXiv:2006.16541}, 2020.

\bibitem[Wightman(2019)]{rw2019timm}
Ross Wightman.
\newblock Pytorch image models.
\newblock \url{https://github.com/rwightman/pytorch-image-models}, 2019.

\bibitem[Wortsman et~al.(2024)Wortsman, Liu, Xiao, Everett, Alemi, Adlam, Co-Reyes, Gur, Kumar, Novak, Pennington, Sohl-Dickstein, Xu, Lee, Gilmer, and Kornblith]{wortsman2024smallscale}
Mitchell Wortsman, Peter~J Liu, Lechao Xiao, Katie~E Everett, Alexander~A Alemi, Ben Adlam, John~D Co-Reyes, Izzeddin Gur, Abhishek Kumar, Roman Novak, Jeffrey Pennington, Jascha Sohl-Dickstein, Kelvin Xu, Jaehoon Lee, Justin Gilmer, and Simon Kornblith.
\newblock Small-scale proxies for large-scale transformer training instabilities.
\newblock In \emph{The Twelfth International Conference on Learning Representations}, 2024.
\newblock URL \url{https://openreview.net/forum?id=d8w0pmvXbZ}.

\bibitem[Yang et~al.(2021)Yang, Hu, Babuschkin, Sidor, Liu, Farhi, Ryder, Pachocki, Chen, and Gao]{Yang21}
Ge~Yang, Edward Hu, Igor Babuschkin, Szymon Sidor, Xiaodong Liu, David Farhi, Nick Ryder, Jakub Pachocki, Weizhu Chen, and Jianfeng Gao.
\newblock Tuning large neural networks via zero-shot hyperparameter transfer.
\newblock In M.~Ranzato, A.~Beygelzimer, Y.~Dauphin, P.S. Liang, and J.~Wortman Vaughan (eds.), \emph{Advances in Neural Information Processing Systems}, volume~34, pp.\  17084--17097. Curran Associates, Inc., 2021.
\newblock URL \url{https://proceedings.neurips.cc/paper_files/paper/2021/file/8df7c2e3c3c3be098ef7b382bd2c37ba-Paper.pdf}.

\bibitem[Zhai et~al.(2022)Zhai, Kolesnikov, Houlsby, and Beyer]{Zhai0HB22}
Xiaohua Zhai, Alexander Kolesnikov, Neil Houlsby, and Lucas Beyer.
\newblock Scaling vision transformers.
\newblock In \emph{{IEEE/CVF} Conference on Computer Vision and Pattern Recognition, {CVPR} 2022, New Orleans, LA, USA, June 18-24, 2022}, pp.\  1204--1213. {IEEE}, 2022.
\newblock \doi{10.1109/CVPR52688.2022.01179}.
\newblock URL \url{https://doi.org/10.1109/CVPR52688.2022.01179}.

\bibitem[Zhang et~al.(2024{\natexlab{a}})Zhang, Morwani, Vyas, Wu, Zou, Ghai, Foster, and Kakade]{zhang2024doescriticalbatchsize}
Hanlin Zhang, Depen Morwani, Nikhil Vyas, Jingfeng Wu, Difan Zou, Udaya Ghai, Dean Foster, and Sham Kakade.
\newblock How does critical batch size scale in pre-training?, 2024{\natexlab{a}}.
\newblock URL \url{https://arxiv.org/abs/2410.21676}.

\bibitem[Zhang et~al.(2024{\natexlab{b}})Zhang, Chen, Ding, Li, Sun, and Luo]{zhang2024transformers}
Yushun Zhang, Congliang Chen, Tian Ding, Ziniu Li, Ruoyu Sun, and Zhi-Quan Luo.
\newblock Why transformers need adam: A hessian perspective, 2024{\natexlab{b}}.

\bibitem[Zhang et~al.(2024{\natexlab{c}})Zhang, Chen, Li, Ding, Wu, Ye, Luo, and Sun]{zhang2024adamminiusefewerlearning}
Yushun Zhang, Congliang Chen, Ziniu Li, Tian Ding, Chenwei Wu, Yinyu Ye, Zhi-Quan Luo, and Ruoyu Sun.
\newblock Adam-mini: Use fewer learning rates to gain more, 2024{\natexlab{c}}.
\newblock URL \url{https://arxiv.org/abs/2406.16793}.

\bibitem[Zheng \& Kwok(2019)Zheng and Kwok]{bagm}
Shuai Zheng and James~T. Kwok.
\newblock Blockwise adaptivity: Faster training and better generalization in deep learning.
\newblock \emph{CoRR}, abs/1905.09899, 2019.
\newblock URL \url{http://arxiv.org/abs/1905.09899}.

\end{thebibliography}
\bibliographystyle{iclr2025_conference}

\newpage
\appendix
\section{Related work}
One closely related work to ours is \citet{wortsman2024smallscale}, which explores the stability of Adam with respect to learning rate. We extend the comparison to other optimizers including SGD, Lion and Adafactor, as well as other hyperparameters including momentum and weight decay.
 
\textbf{Optimizers:} SGD \citep{robbinsmonro51} had been the workhorse optimizer for deep learning until 2015, when Adam \citep{kingma15} was introduced. Adam is a diagonal preconditioning algorithm that maintains a per-parameter learning rate. Over time, coarser variants of Adam have been proposed, which do not explicitly maintain a learning rate per parameter. Adafactor \citep{ShazeerS18, Zhai0HB22} maintains a rank-1 approximation of the preconditioner matrix of Adam.  Previous works have also explored Signum \citep{pmlr-v80-bernstein18a, bernstein2018signsgd} and have observed its benefits in terms of communication efficiency and fault tolerance. Other works have also explored the similarity of Adam with variants of Signum \citep{balles2018dissecting}, and recently, a close variant of Signum, called Lion \citep{chen2023symbolic}, was discovered using symbolic search over algorithms. Some other optimizers that have recently gained increasing attention from the community include Shampoo~\citep{shampoo} and Sophia \citep{liu2024sophia}.

Similar to us, prior work has explicitly studied optimizers comparisons across different domains and architectures. For instance, \citet{schmidt2021descending} performs a sweep over optimizers, schedulers, and seeds over various vision tasks. As mentioned in the introduction, \citet{kaddour2024no} compares optimizers for masked language modeling and at a fixed model scale. Other works have called to attention better benchmarking practices for optimizers~\citep{schneiderdeepobs, sivaprasad2020optimizer, bartz2020benchmarking}.

\textbf{Adam and Signum:}  Many works have explored the relationship between Adam and variants of Signum \citep{balles2018dissecting, balles2020geometry, kunstner2023noise} and empirically demonstrated that Signum (or its close variants) generally performs comparably to Adam. \citet{balles2020geometry} also argued that signSGD generally performs better when the Hessian is close to diagonal, however, it is unclear if this holds for practical settings. \citet{kunstner2023noise} recently demonstrated that Adam and a close variant of Signum exhibit similar performance on a variety of datasets including WikiText-2 \citep{merity2017pointer} and SQuAD \citep{rajpurkar2016squad}. However, in contrast with our work, all of these are restricted to the setting of vision or masked language modeling, and generally do not sweep over multiple hyperparameters.

\textbf{Layerwise or blockwise Adam:} We study Adalayer, a layerwise version of Adam. This is a special case of the BAGM optimizer~\citep{bagm}, specifically BAGM B.1. Similar algorithms have also been studied by previous works \citep{ginsburg2019stochastic,grafting,pmlr-v139-liu21c, zhang2024adamminiusefewerlearning}. In particular, concurrent to our work, \citet{zhang2024adamminiusefewerlearning} propose an algorithm termed Adam-mini, which closely tracks a modified version of Adalayer (called Adalayer*), and demonstrate comparable performance to AdamW. Note that, in our work, Adalayer* is introduced to understand the role played by preconditioning in Adam, and we do not specifically focus on the final performance.  \citet{zhang2024transformers} empirically study the Hessian spectrum of transformers at initialization and find it to be more heterogeneous across layers as compared to ResNets. They argue that this heterogeneity is evidence towards the importance of Adam in training transformers. In contrast our results (Section~\ref{subsec:sgd+novograd}) show that Adam's preconditioning is particularly important for the last layer and LayerNorm parameters to achieve performance and learning rate stability.

\textbf{Other related works:} For vision transformers, in the fine-tuning phase, \citet{kumar2024how} show that using SGD with frozen embedding parameters leads to competitive performance with Adam. \citet{jelassi2022dissecting} explore the similarity between Adam and normalized gradient descent \citep{nesterov2004introductory} and show that normalized gradient descent on GANs does not suffer from mode collapse, while SGD does. \citet{jiang2023how} empirically demonstrate that Adam steers the parameter trajectory towards better-conditioned regions than SGD. \citet{pan2022toward} also show that the parameter trajectory of Adam exhibits much higher directional smoothness than that of SGD. \citet{ahn2024linear} show that the performance gap between Adam and SGD exacerbates with depth of the network. In a similar vein to us, \citet{kunstner2024heavy} show that Adam is less sensitive than gradient descent to class-imbalance present in language tasks; we provide further evidence for the importance of preconditioning the last layer, as well as the LayerNorm parameters.

\section{Additional Main Sweep Results}
\label{app:details}

\begin{table}[h!]
\centering
\begin{tabular}{|c|c|}
\hline
\textbf{Optimizer} & \textbf{Optimal Learning Rate} \\ \hline
Adam & 3.16e-3 (150m), 1e-3 (300m), 1e-3 (600m), 1e-3 (1.2b) \\ \hline
Adafactor & 3.16e-3 (150m), 1e-3 (300m), 1e-3 (600m), 1e-3 (1.2b) \\ \hline
Lion & 3.16e-4 (150m), 3.16e-4 (300m), 3.16e-4 (600m), 1e-4 (1.2b) \\ \hline
Signum & 3.16e-4 (150m), 3.16e-4 (300m), 3.16e-4 (600m), 3.16e-4 (1.2b) \\ \hline
\end{tabular}
\caption{Optimal learning rates for various optimizers from Figure~\ref{fig:main}.}
\label{tab:optimizers}
\end{table}

For our ablations in Figure~\ref{fig:main}, we report on the optimal learning rate found for each optimizer in Table~\ref{tab:optimizers}. In general, we find that the optimal learning rate for Adam and Adafactor are similar, with the optimal learning rate of Lion and Signum an order of magnitude smaller.

\section{Adalayer}
\label{app:adalayer}
As mentioned in Section~\ref{sec:novograd_investigation}, to investigate the role of preconditioning on language models for optimizers like Adam, we introduce the Adalayer optimizer for ease of analysis. In this section, we first establish the performance and stability of Adalayer as a reasonable proxy for Adam by making a modification to how Adalayer treats the last layer of the network.

\begin{figure}[h]
    \centering
    \includegraphics[width=0.5\textwidth]{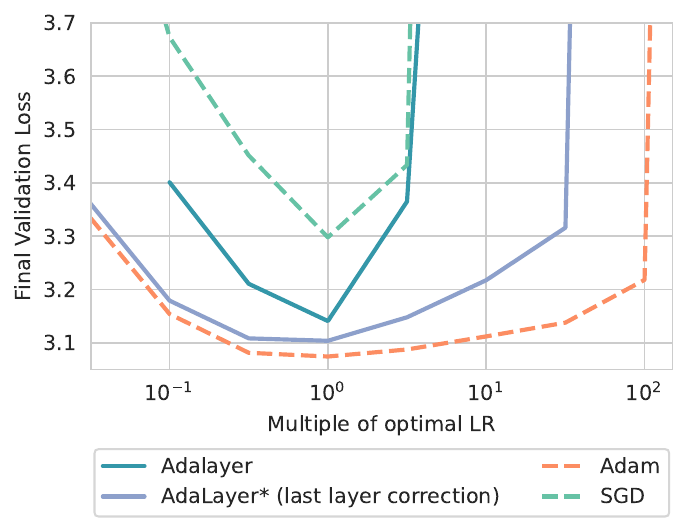}
    \caption{Modifying Adalayer with the last layer correction improves performance and stability across learning rates.}
    \label{fig:novograd_last_correction}
\end{figure}

In Figure~\ref{fig:novograd_last_correction} we study the behavior of Adalayer across learning rates. To preserve the correspondence with Adam we fix other hyperparameters to be the same: $\beta_1 = 0.9$, $\beta_2 = 0.95$ and $\epsilon = 1e-15$. We find that Adalayer has better performance than SGD, but it performs worse than Adam and also lacks Adam's stability across learning rates. The major difference between Adam and Adalayer is the preconditioning done by Adam \textit{within} a layer. Intuitively, this preconditioning will have large effects in layers where we expect different weights within a layer to have different gradient scales. The first candidate for such a layer is the \emph{last layer}, since different tokens have widely different frequencies leading to different gradient scales. To test this hypothesis, we run a corrected version\footnote{We note that this reasoning also applies to the first layer, but in our ablations applying this correction the first layer did not make a significant difference.} of Adalayer where we treat the set of weights feeding into a logit as a separate block. We henceforth refer to Adalayer with this correction as \textbf{Adalayer*}. This is plotted in Figure~\ref{fig:novograd_last_correction} and we observe that Adalayer* almost recovers the performance as well as a large fraction of the stability of Adam. 

\section{Additional experiments: SGD + adaptive variants (Adalayer*, Adafactor)}
\label{app:sgd_adalayer}
\begin{figure}[h]
    \centering
    \includegraphics[width=0.5\textwidth]{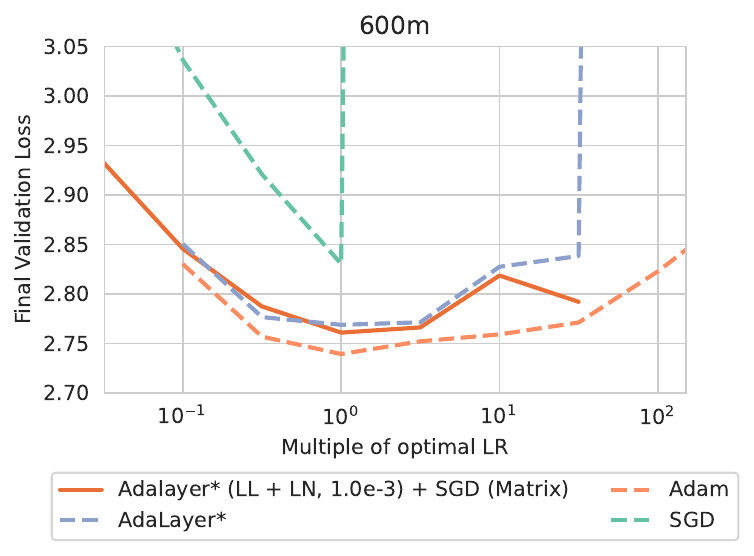}
    \caption{As in Section~\ref{subsec:sgd+novograd}, we train 300m (\textbf{Left}) and 600m models (\textbf{Right}) with Adalayer* on the last layer and LayerNorm parameters, and train the remaining model parameters with SGD. We see that performance and stability continues to match that of Adalayer* or outperform Adalayer* even at these larger scales..}
    \label{fig:sgd_novograd_600}
\end{figure}

In this section, we report additional experiments involving training language models with SGD on a fraction of the models' parameters and an adaptive optimizer on the remaining parameters. Firstly, we show that our results from Section~\ref{subsec:sgd+novograd} hold even when training 600m parameter models with Adalayer* applied only on the last layer and LayerNorm parameters in Figure~\ref{fig:sgd_novograd_600}.

\begin{figure}[h]
    \centering
    \includegraphics[width=0.45\textwidth]{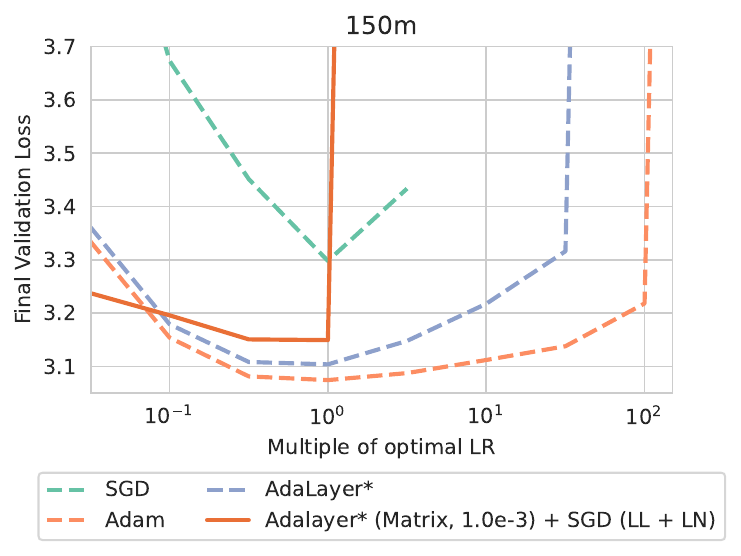}
    \caption{We train 150m models using Adalayer* on the matrix layers with a fixed learning rate of $1e-3$ and using SGD on the last layer and LayerNorm parameters. Compared to the results in Figure \ref{fig:sgd+novograd}, we do not recover the same stability nor do we reach the optimal performance of Adalayer*.}
    \label{fig:sgd_novograd_matrix_only}
\end{figure}

We also provide further ablations supporting our claim that the largest impact of the adaptivity of Adalayer* is concentrated on the last layer and LayerNorm parameters. Firstly, we train 150m models using Adalayer* on only the matrix parameters, while training the last layer and LayerNorm parameters with SGD. In Figure~\ref{fig:sgd_novograd_matrix_only}, we see performance improves relative to SGD but we see similar instability at larger learning rates.

\begin{figure}[h]
    \centering
    \includegraphics[width=0.45\textwidth]{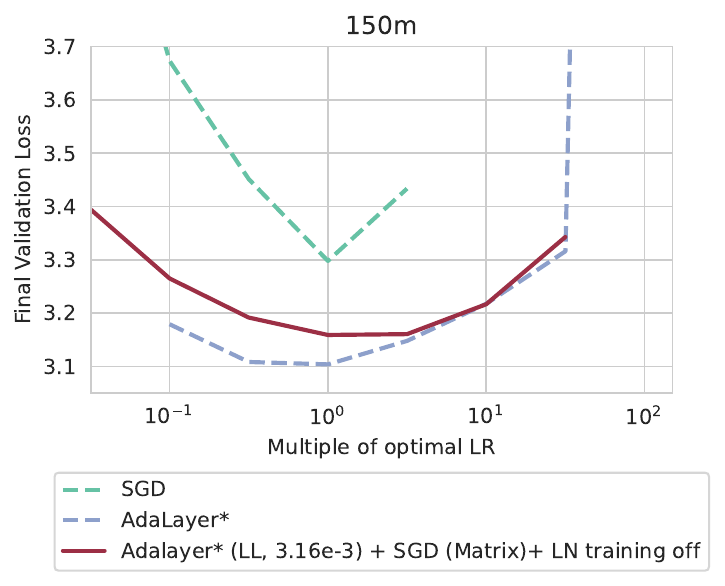}
    \includegraphics[width=0.45\textwidth]{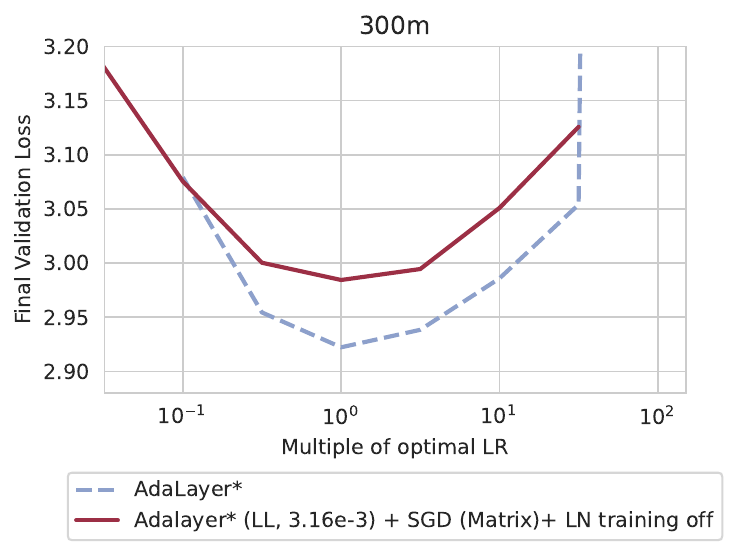}
    \caption{Training 150m (\textbf{Left}) and 300m (\textbf{Right}) models using Adalayer* on the last layer with a fixed learning rate of $3.16e-3$ and using SGD on other matrix blocks, while turning off the option to train LayerNorm parameters. We see that while the performance and stability is improved compared to SGD, it is still not as performant as Adalayer*. This indicates a degree of importance of training LayerNorm parameters for these models.}
    \label{fig:sgd_novograd_ln_off}
\end{figure}

Secondly, given that the effective learning rates of the LayerNorm blocks were observed to be small in Figure~\ref{fig:eff-lr-all} (\textbf{Right}), it is reasonable to ask whether training the LayerNorm parameters is necessary at all; in Figure~\ref{fig:sgd_novograd_ln_off}, we show results for training 150m and 300m models using Adalayer* only on the last layer, using SGD on all other matrix blocks, and turning off training for the LayerNorm parameters. This indeed yields greater stability in comparison to Figure~\ref{fig:sgd+novograd} (\textbf{Left}) but does not fully recover the performance of Adalayer*, indicating that training LayerNorm parameters helps with performance, which seems more pronounced in the larger model.

\label{app:sgd_variants}

\begin{figure}[h]
    \centering
    \includegraphics[width=0.45\textwidth]{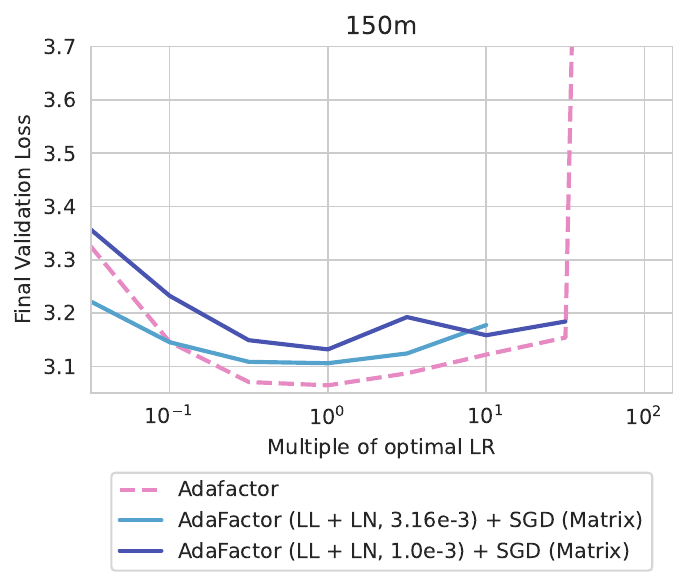}
    \includegraphics[width=0.45\textwidth]{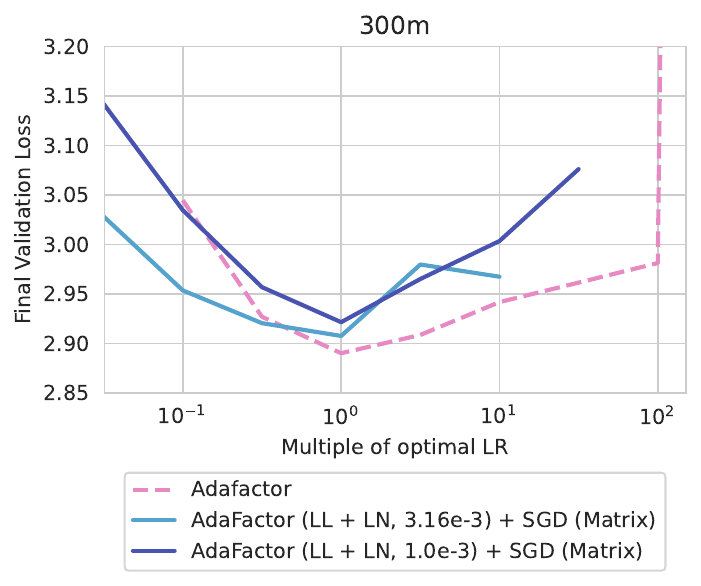}
    \caption{Training 150m (\textbf{Left}) and 300m (\textbf{Right}) models using Adafactor on the last layer with a fixed learning rate and using SGD on other matrix blocks. We see that performance and stability is comparable to Adafactor, but does not exceed it, particularly at higher learning rates.}
    \label{fig:sgd_adafactor}
\end{figure}

We saw in Figure~\ref{fig:sgd+novograd} (\textbf{Middle, Right}) that using Adalayer* on only the last layer and LayerNorm parameters sufficed to recover or exceed the performance of Adalayer*. In Figure~\ref{fig:sgd_adafactor}, we report a learning rate sweep over the analogous experiment but using Adafactor on the last layer and LayerNorm parameters with a fixed learning rate. For the 150m model, using a learning rate of $3.16e-3$ with Adafactor yielded better performance than Adafactor for low learning rates, and is comparable in terms of performance and stability. For the 300m model, the difference between Adafactor and our `hybrid' optimizer is more distinct at higher learning rates for fixed Adafactor learning rate $1.0e-3$ and $3.16e-3$, but is comparable until the peak validation loss.

\section{Additional Experiments: AdaSGD + Adalayer*}
\label{app:adasgd}

The previous experiments on SGD + Adalayer* show that although using a fixed learning rate for the matrix parameters largely recovers performance of full Adalayer*, there remains a small gap to full Adalayer* or Adam performance. Can we recover what is remaining? We further investigate the importance of adaptivity at different levels of granularity by training the matrix parameters with AdaSGD~\citep{wang2020adasgd} instead of SGD. AdaSGD still uses a global learning rate but allows this scalar to be adaptive across training--- in other words, this is a ``global'' version of Adalayer* where all matrix parameters are treated as one block. 

\begin{figure}[h]
    \centering
    \includegraphics[width=0.65\textwidth]{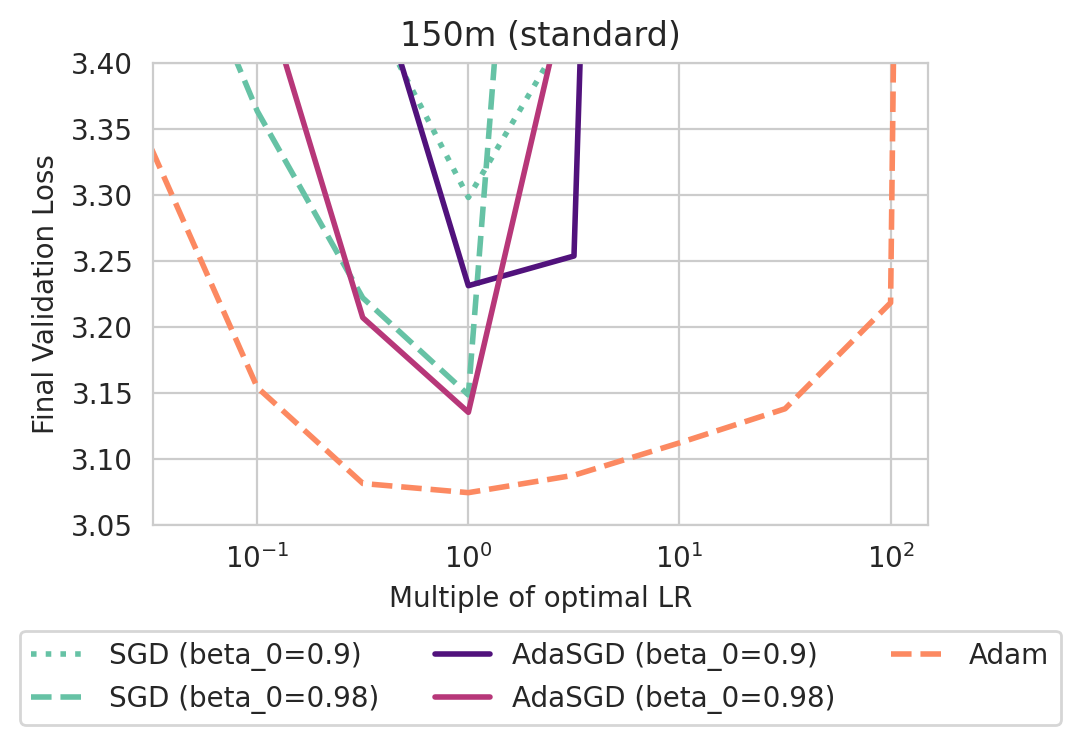}
    
    \caption{Learning rate sweep for AdaSGD with two $\beta_0$ values. We observe that AdaSGD performs marginally better than SGD but still lacks learning rate stability.}
    \label{fig:adasgd}
\end{figure}

In Figure~\ref{fig:adasgd} we conduct an analogous learning rate sweep for AdaSGD for $\beta_0 \in \{0.9, 0.98\}$ as we did in Section~\ref{sec:ablations}. The performance of AdaSGD is marginally better than SGD, but still suffers from learning rate instability; this is consistent with our previous findings in Figure~\ref{fig:eff-lr-all}, where we saw that certain parameters had effective learning rates which were multiple orders of magnitude different from other parameters, and thus a single adaptive learning rate across all parameters likely would not suffice for achieving performance and stability.

\begin{figure}[h]
    \centering
    \includegraphics[width=0.65\textwidth]{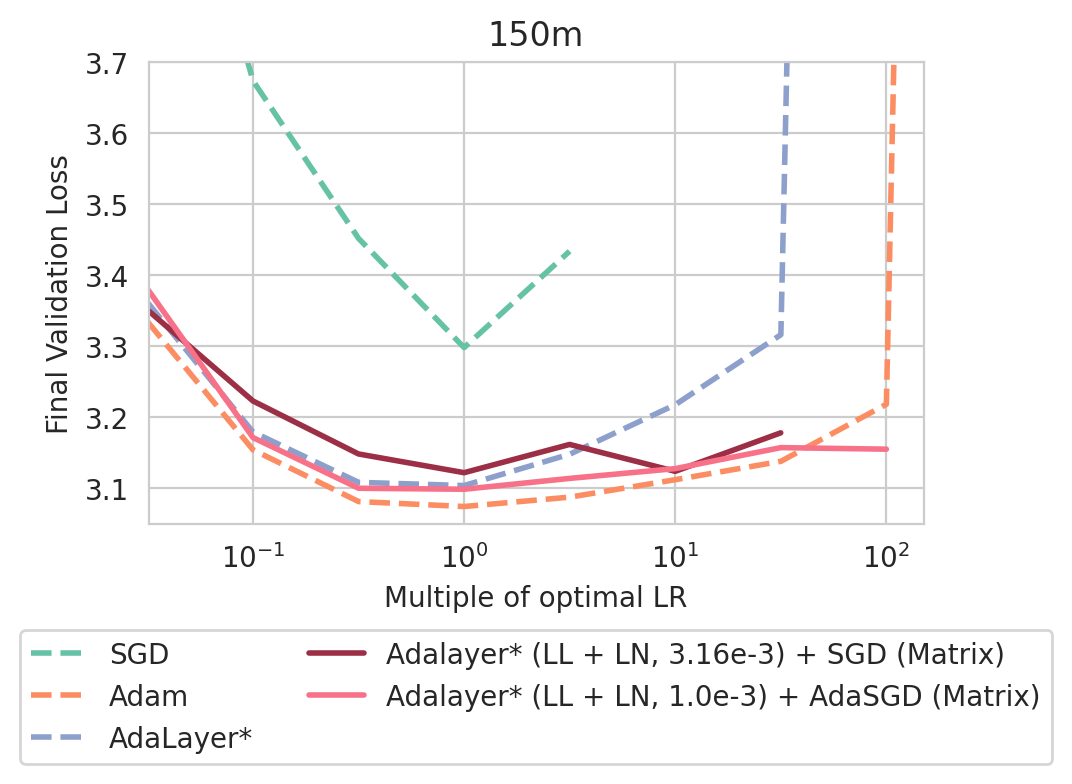}
    
    \caption{When replacing SGD with AdaSGD on the matrix parameters and using Adalayer* to train the last layer and LayerNorm parameters, the performance and stability has improved to match that of Adalayer*, but a gap still remains between Adam performance even with an adaptive learning rate across matrix parameters.}
    \label{fig:adasgd_adalayer}
\end{figure}

In Figure~\ref{fig:adasgd_adalayer} we replace SGD with AdaSGD on the matrix parameters and use Adalayer* to train the last layer and LayerNorm parameters. The performance now essentially matches that of full Adalayer* compared to SGD + Adalayer*, showing that the single adaptive learning rate on matrix parameters is largely sufficient to recover Adalayer* performance. However, a gap still remains between Adam performance even with an adaptive learning rate across matrix parameters; it is likely that this small gap is attributed to the need for increased adaptivity granularity for certain matrix parameters.

\section{Additional experiments: freezing Adalayer learning rate ratios}
\label{app:novograd_freeze}
\begin{figure}[!ht]
    \centering
    \includegraphics[width=0.45\textwidth]{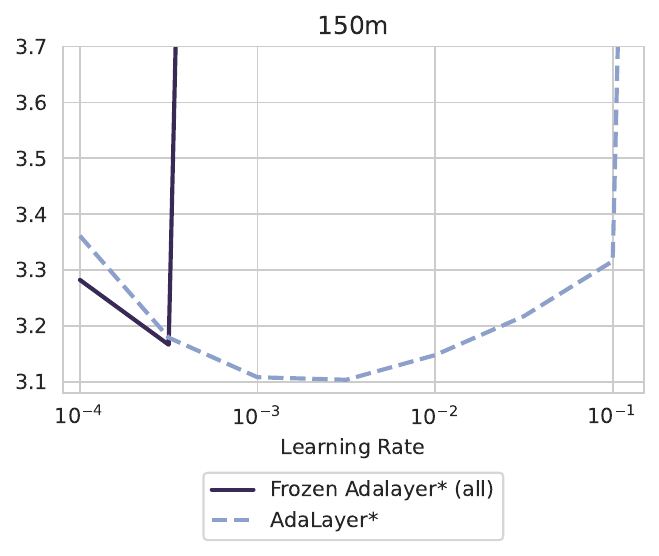}
    \includegraphics[width=0.45\textwidth]{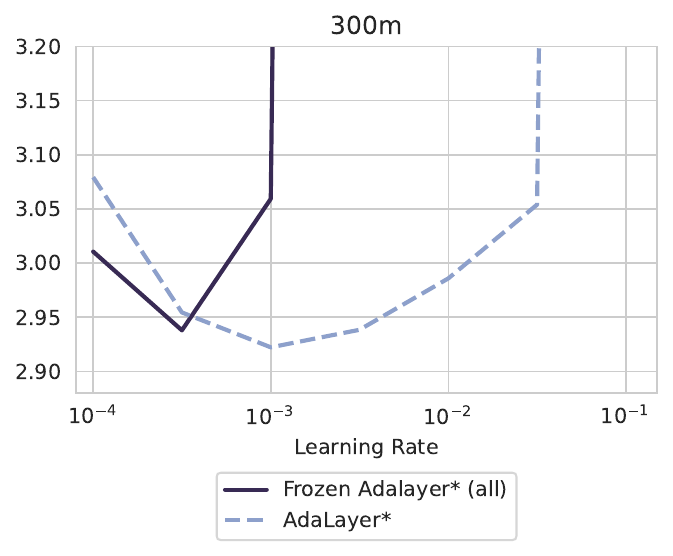}
    \caption{Training 150m (\textbf{Left}) and 300m (\textbf{Right}) models using fixed Adalayer* learning rate ratios from initialization for \emph{all layers}. We observe this quickly diverges, achieving neither peak performance nor stability.}
    \label{fig:novograd_freeze_all}
\end{figure}

\begin{figure}[!ht]
    \centering
    \includegraphics[width=0.45\textwidth]{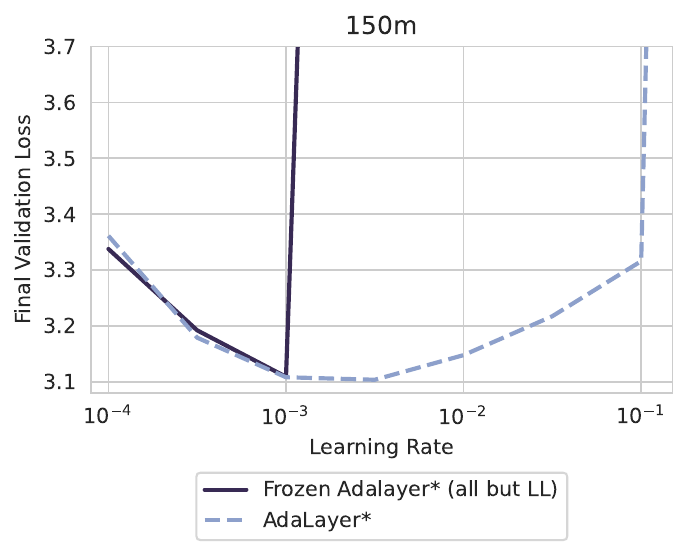}
    \includegraphics[width=0.45\textwidth]{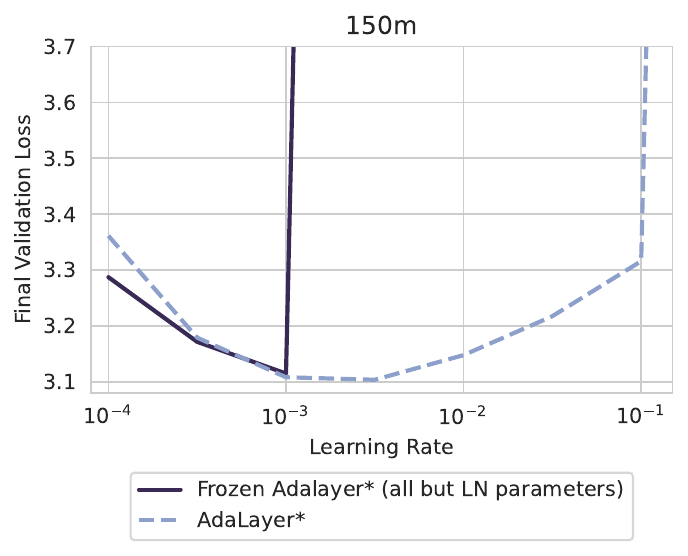}
    \caption{Training 150m models using fixed Adalayer* learning rate ratios from initialization while either excluding only the last layer (\textbf{Left}) or excluding only the LayerNorm parameters (\textbf{Right}). We observe both modifications reach peak performance but fails to be stable at higher learning rates.}
    \label{fig:novograd_freeze_one}
\end{figure}

\begin{figure}[!ht]
    \centering
    \includegraphics[width=0.45\textwidth]{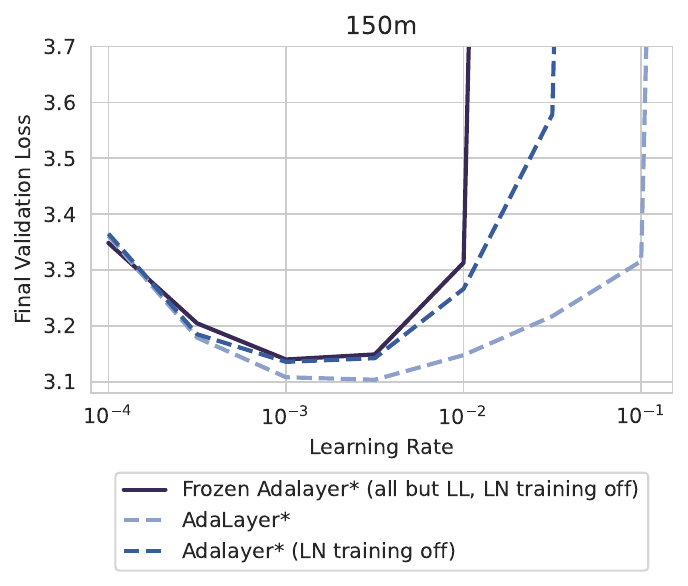}
    \includegraphics[width=0.45\textwidth]{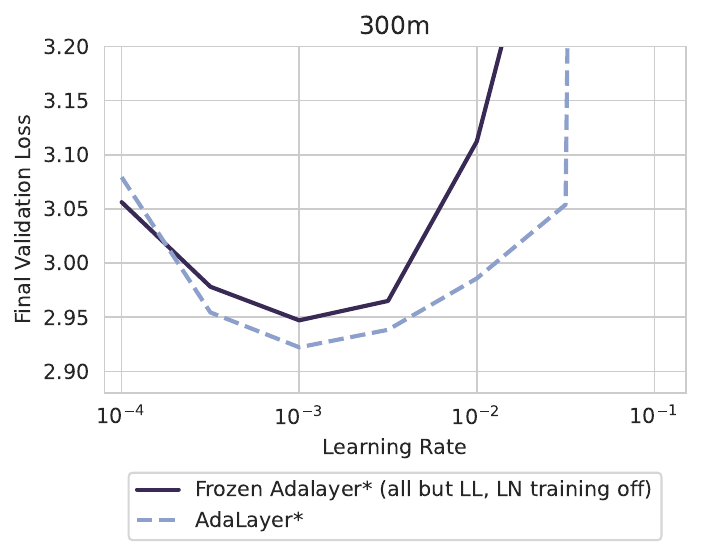}
    \caption{Training 150m (\textbf{Left}) and 300m (\textbf{Right}) models using fixed Adalayer* learning rate ratios from initialization, while letting the last layer continue to update, and turning LayerNorm training off. Stability across learning rates has improved but is less performant; for the 150m model we also plot the sweep for regular Adalayer* with LayerNorm training off, and we see that it is worse in performance compared to Adalayer* with LayerNorm training.}
    \label{fig:novograd_freeze_ln_off}
\end{figure}

In this section, we report additional experiments exploring whether the results involving frozen Adalayer* in Section~\ref{subsec:sgd+novograd} need both last layer and LayerNorm adaptivity. We show that this is indeed the case by conducting the same sweep for frozen Adalayer* while trying to also freeze the learning rate ratios for last layer or LayerNorm parameters as well. In Figure~\ref{fig:novograd_freeze_all}, we show that fixing initialized learning rate ratios for \emph{all} layers does not reach peak performance of Adalayer*, nor does it exhibit stability. In Figure~\ref{fig:novograd_freeze_one}, we show that either continuing to update the LayerNorm parameters or the last layer parameters can achieve the peak performance of Adalayer* but is still unstable. Finally, we show results for turning off LayerNorm training while fixing learning rate ratios (with the exception of the last layer) in Figure~\ref{fig:novograd_freeze_ln_off}. We conclude that it is necessary to maintain adaptivity for \emph{both} the last layer and LayerNorm parameters, but understanding why the fixed ratios do not suffice would be an interesting question for future work.

\clearpage

\section{Sophia}
\label{app:sophia}

In this section, we compare Sophia~\citep{liu2024sophia} to Signum. Note that Signum is a special case of Sophia, achieved by setting $\rho = 0$. We find that Sophia does not outperform Signum. No significant change in performance was observed when transferring the hyperparameters suggested by \citet{liu2024sophia} (eg. $\beta_1$, $\beta_2$, $\varepsilon$, weight decay), nor when additionally scaling attention by the inverse of layer index which was used in the original Sophia implementation.

\begin{figure}[h]
    \centering
    \includegraphics[width=0.65\textwidth]{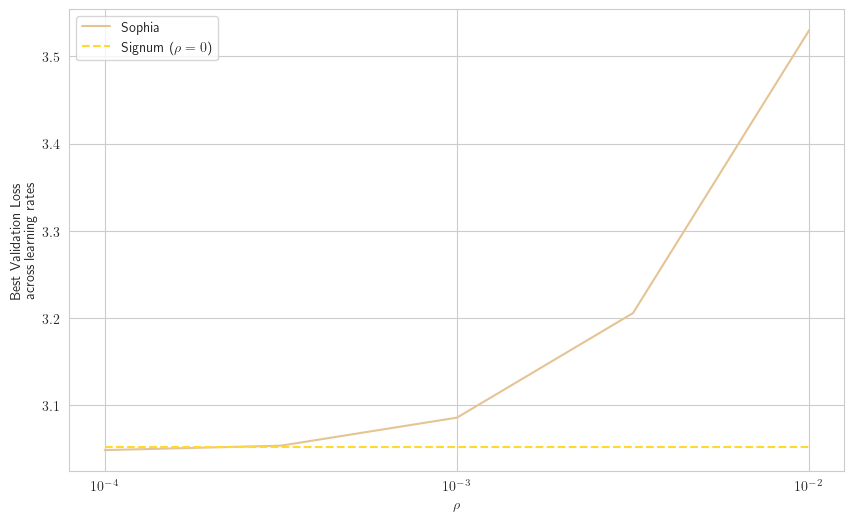}
    
    \caption{Comparing Sophia~\citep{liu2024sophia} and Signum for the 150M model in our default setup.}
    \label{fig:sophia}
\end{figure}

\end{document}